\documentclass[acmsmall]{acmart}  
\AtBeginDocument{%
  \providecommand\BibTeX{{%
    \normalfont B\kern-0.5em{\scshape i\kern-0.25em b}\kern-0.8em\TeX}}}

\usepackage{graphicx}
\usepackage{amsfonts}
\usepackage{amsmath,bm,bbm}
\usepackage{url}
\usepackage{endnotes}
\usepackage[ruled,boxed,linesnumbered]{algorithm2e}
\usepackage{booktabs}
\usepackage{color}
\usepackage{colortbl}
\usepackage{subfigure}
\usepackage{tabularx}
\usepackage{multirow}
\usepackage{verbatim}
\usepackage{pifont}

\usepackage{xcolor}

\newcommand{\argmin}[1]{\arg\underset{#1}{\min}\; }

\definecolor{darkgreen}{RGB}{50,150,50}

\newcommand{\etc}{\emph{etc.~}}
\newcommand{\ie}{\emph{i.e.,~}}
\newcommand{\etal}{\emph{et al.~}}
\newcommand{\eg}{\emph{e.g.,~}}

\newcommand{\specialcell}[2][l]{%
  \begin{tabular}[#1]{@{}l@{}}#2\end{tabular}}

\newcommand{\PreserveBackslash}[1]{\let\temp=\\#1\let\\=\temp}
\newcolumntype{C}[1]{>{\PreserveBackslash\centering}p{#1}}
\newcolumntype{R}[1]{>{\PreserveBackslash\raggedleft}p{#1}}
\newcolumntype{L}[1]{>{\PreserveBackslash\raggedright}p{#1}}

\newfont{\mycrnotice}{ptmr8t at 7pt}
\newfont{\myconfname}{ptmri8t at 7pt}
\setcopyright{acmcopyright}
\copyrightyear{2021}
\acmYear{2021}

\acmJournal{TOMM}

\begin{document}

\title{Unsupervised Domain Expansion for Visual Categorization}







\author{Jie Wang}
\authornote{Equal contribution}
\affiliation{\institution{Key Lab of DEKE, Renmin University of China}}

\author{Kaibin Tian}
\authornotemark[1]
\affiliation{\institution{Key Lab of DEKE, Renmin University of China}}

\author{Dayong Ding}
\affiliation{\institution{Vistel AI Lab, Visionary Intelligence Ltd. Beijing}}

\author{Gang Yang}
\affiliation{\institution{School of Information, Renmin University of China}}

\author{Xirong Li}
\authornote{Corresponding author: Xirong Li (xirong@ruc.edu.cn)}
\affiliation{\institution{Key Lab of DEKE, Renmin University of China}}

\renewcommand{\shortauthors}{Wang et al.}

\begin{abstract}
Expanding visual categorization into a novel domain without the need of extra annotation has been a long-term interest for multimedia intelligence. Previously, this challenge has been  approached by unsupervised domain adaptation (UDA). Given labeled data from a source domain and unlabeled data from a target domain, UDA seeks for a deep representation that is both discriminative and domain-invariant. While UDA focuses on the target domain, we argue that the performance on both source and target domains matters, as in practice which domain a test example comes from is unknown. In this paper we extend UDA by proposing a new task called unsupervised domain expansion (UDE), which aims to adapt a deep model for the target domain with its unlabeled data, meanwhile maintaining the model's performance on the source domain. We propose Knowledge Distillation Domain Expansion (KDDE) as a general method for the UDE task. Its domain-adaptation module can be instantiated with any existing model. We develop a knowledge distillation based learning mechanism, enabling KDDE to optimize a single objective wherein the source and target domains are equally treated. Extensive experiments on two major benchmarks, \ie Office-Home and DomainNet, show that KDDE compares favorably against four competitive baselines, \ie DDC, DANN, DAAN, and CDAN, for both UDA and UDE tasks. Our study also reveals that the current UDA models improve their performance on the target domain at the cost of noticeable performance loss on the source domain.
\end{abstract}



\maketitle

\section{Introduction} \label{sec:intro}

\begin{figure}[tbh!]
  \setlength{\abovecaptionskip}{5pt}
  \centering
  \includegraphics[width=\textwidth]{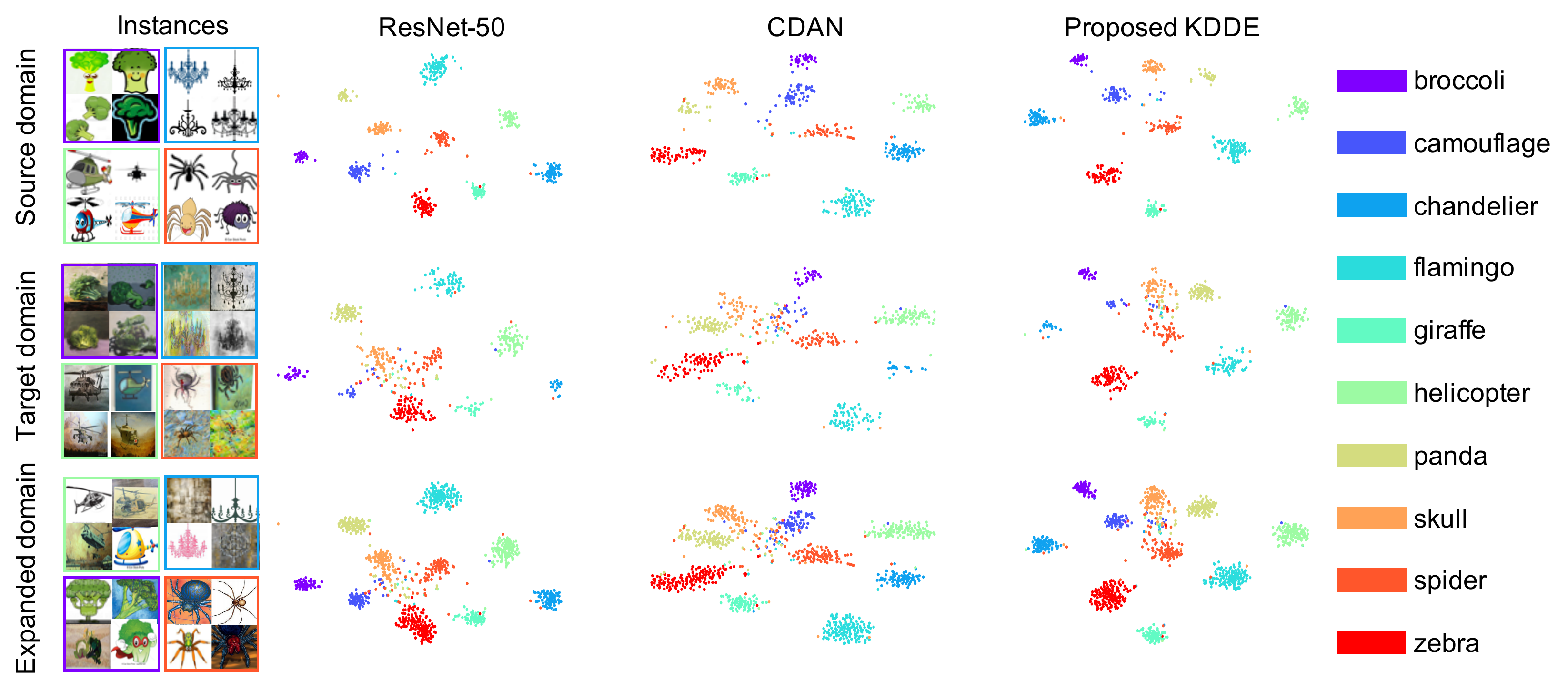}
  \caption{\textbf{Visualization of deep feature spaces obtained by three models}, that is, \emph{ResNet-50} fully trained on a labeled source domain, \emph{CDAN}~\cite{nips18-cdan} as a state-of-the-art domain adaptation model and the proposed Knowledge Distillation Domain Expansion (\emph{KDDE}) that adaptively learns from both ResNet-50 and CDAN. Across the source, target and expanded domains, intra-class data points tend to stay closer while inter-class data points are more distant in the feature space of KDDE. With no need of extra labeled data, KDDE effectively expands the applicable domain of visual classifiers.}
  \label{fig:tsne}
\end{figure}

It has been recognized early by the multimedia community that visual classifiers trained on a specific domain do not necessarily perform well on a distinct domain~\cite{mm07-da,icip08-da,cvpr09-da}. Even for the same concept, \eg \textit{helicopter}, changing from one domain to another, \eg \textit{clipart} $\rightarrow$ \textit{painting}, would result in significant discrepancy in visual appearance, see Fig. \ref{fig:tsne}. When such discrepancy is propagated to the feature space wherein classification is performed, performance degeneration occurs. Note that the advance of deep learning does not alleviate the problem. Rather, due to its ``super'' learning ability, deep representations tend to be dataset biased~\cite{aaai18-datasetbias}.

In order to improve the generalization ability of a deep visual classifier without the need of extra annotation, deep unsupervised domain adaptation (UDA) has been actively studied~\cite{ddc,eccvw16-deep-coral,nips18-cdan,mm19-jada}. Given the availability of labeled data from a source domain and unlabeled data from a target domain, UDA seeks for a deep representation that is both discriminative and domain-invariant. In the seminal work by Tzeng \etal~\cite{ddc}, a deep convolutional neural network termed Deep Domain Confusion (DDC) is developed to simultaneously minimize the classification loss and a domain discrepancy loss computed in terms of first-order statistics of the deep features from the two domains. Follow-ups improve DDC in varied aspects, including Deep CORAL~\cite{eccvw16-deep-coral} that replaces first-order statistics by second-order statistics, JAN~\cite{icml17-jan} that measures domain discrepancy on multiple task-specific layers, and DANN~\cite{jmlr16-dann} and CDAN~\cite{nips18-cdan} that reduce domain discrepancy by adversarial learning, to name just a few.


While the above efforts have accomplished well for the UDA task, how they perform in the original source domain is mostly unreported, to the best of our knowledge. The absence of performance evaluation on the source domain rises an important question: \emph{is a domain-adapted model indeed domain-invariant?} A follow-up question is \emph{whether the performance gain for the target domain is obtained at the cost of significant performance loss in the source domain?} These two questions are so far overlooked, as existing works on UDA typically assume to know which is the target distribution to tackle. In synthetic-to-real UDA \cite{zou2018unsupervised,2020Unsupervised}, for instance, one treats the synthetic data as the source domain and thus only the real dataset performance matters in the end. However,  
we argue that the performance on both source and target domains matters, as in practice which domain a test example comes from can be unknown. In this paper we extend UDA by proposing a new task, which aims to adapt a deep model for the target domain with its unlabeled data, meanwhile maintaining the model's performance on the source domain. As this factually expands the model's applicable domain, we coin the new task \emph{unsupervised domain expansion} (UDE). 

\begin{figure}[tbh!]
  \setlength{\abovecaptionskip}{5pt}
  \centering
  \includegraphics[width=\columnwidth]{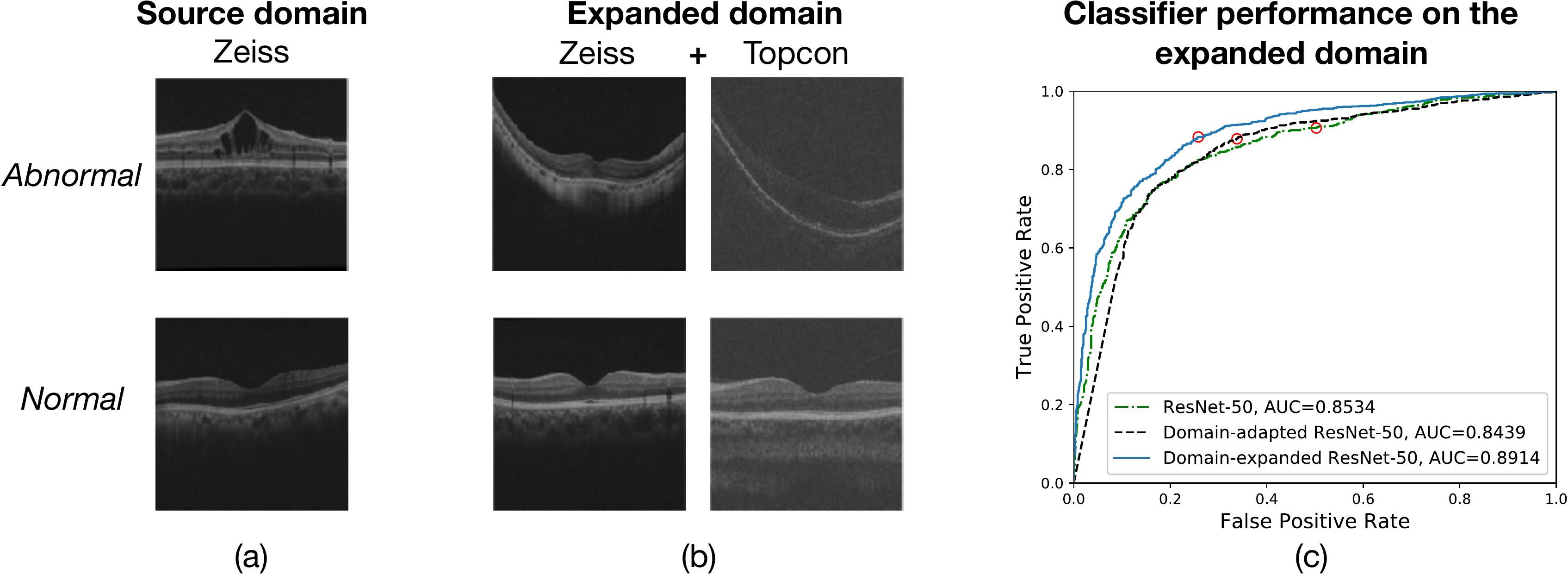}
  \caption{\textbf{Abnormal OCT image recognition as a showcase of Unsupervised Domain Expansion (UDE)}. OCT images taken by distinct devices, \eg (a) \emph{Zeiss Cirrus OCT} and (b) \emph{Topcon 2000FA OCT}, often differ noticeably in their visual appearance. A domain-adapted classifier, departing from the source domain (\emph{Zeiss}), emphasizes its performance on the target domain (\emph{Topcon}). In contrast, a domain-expanded classifier aims for the overall performance of both source and target domains, which is essential for real-world applications. Red circles on (c) ROC curves of the three classifiers indicate their operating points using the default cutoff of 0.5. See Section \ref{sec:exp} for detailed experiments.}
  \label{fig:oct-showcase}
\end{figure}

UDE targets at an expanded domain consisting of test examples from the source and target domains both. So it handles with ease the situation when the domain of a test example is unknown. Such a situation is not uncommon. Consider the medical field for instance. Optical Coherence Tomography (OCT) images, as an important means for ophthalmologists to assess retinal conditions, are actively used for automated retinal screening and referral recommendation~\cite{de2018clinically,kermany2018identifying,miccai19-amd}. As a matter of fact, an eye center is often equipped with multiple types of OCT devices made by distinct manufacturers, let alone multiple eye centers. Meanwhile, OCT images taken by distinct devices can show noticeable discrepancy in their visual appearance, see Fig. \ref{fig:oct-showcase}. Even though UDA improves the generalization ability of a model trained on samples collected from one device for another device, domain adaptation per device is economically unaffordable. With the goal to optimize performance for the expanded domain with a single model, UDE is essential for real-world medical image classification. What is more, by simultaneously evaluating on both source and target domains, UDE provides a direct measure of to what extent a resultant model is domain-invariant, which is mostly missing in the rich literature of UDA.



Although UDA models have considered the performance of the source domain in their learning process, the two objectives to be optimized, \ie \textit{discriminability} and \textit{domain-invariance}, are not always consistent. Such a property makes the existing models difficult to maximize their performance on the expanded source + target domain. While existing models such as DDC and CDAN have a trade-off parameter to balance the two objectives, tunning such a parameter was among our early yet unsuccessful efforts, see Section \ref{sssec:lambda-effect}. A new method for UDE is thus in demand. 

\begin{figure}[tbh!]
  \setlength{\abovecaptionskip}{5pt}
  \centering
  \includegraphics[width=0.9\columnwidth]{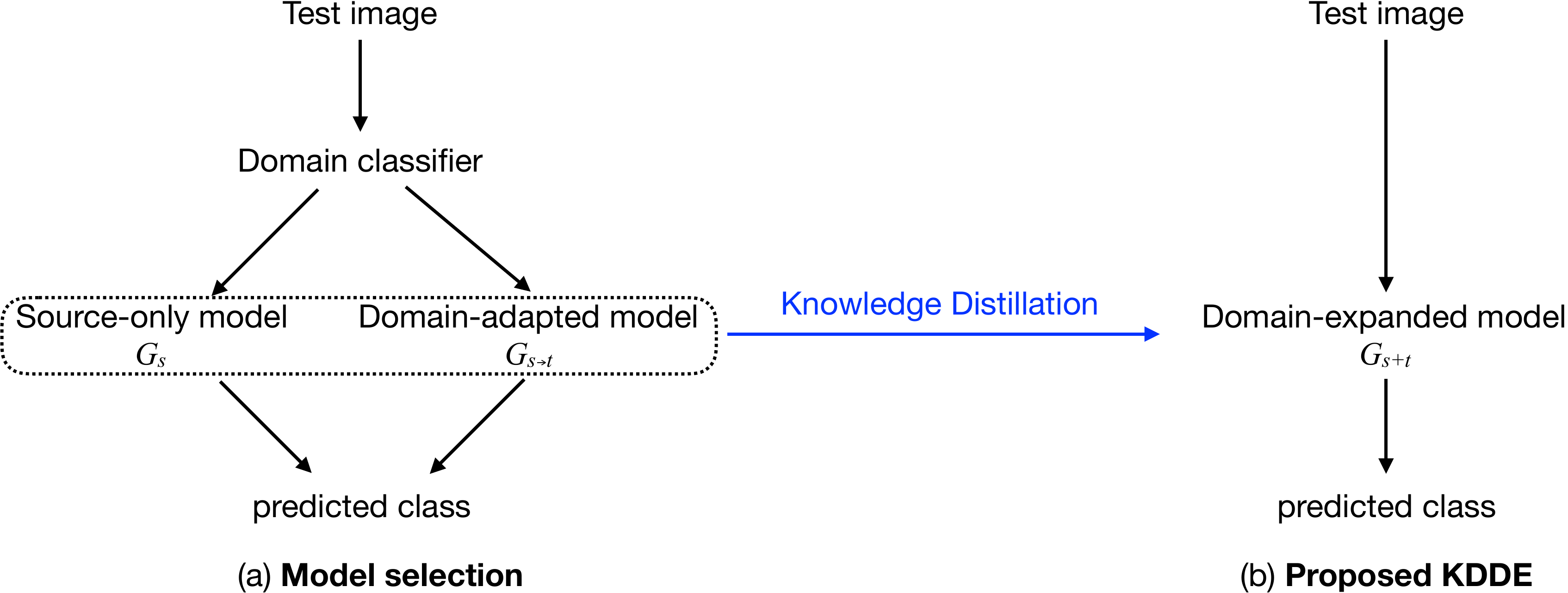}
  \caption{\textbf{Conceptual illustration of two solutions for UDE}. (a) \textbf{Model selection}: A straightforward solution that deploys a source-only model $G_s$ for the source domain and a domain-adapted model $G_{s\rightarrow t}$ for  the target domain, and then chooses which one to use conditioned on the output of a domain classifier. (b) \textbf{Our solution}: Develop a single yet domain-expanded model $G_{s+t}$ by distilling dark knowledge from both $G_s$ and $G_{s\rightarrow t}$.}
  \label{fig:conceptual}
\end{figure}

A straightforward solution for the UDE task is to train a domain classifier to automatically determine which domain a specific test image belongs to. Accordingly, if the test image is deemed to be from the source (or target) domain, a model trained on the source domain (or another model adapted w.r.t. the target domain) is selected to handle the image, see Fig. \ref{fig:conceptual}~(a). Such a solution is cumbersome as it needs to deploy three classifiers at the inference stage. Moreover, misclassification by the domain classifier will make the test image assigned to an inappropriate model. By contrast, we aim for a single domain-expanded model that handle test images from both domains in an unbiased manner, as illustrated in Fig. \ref{fig:conceptual}~(b). To that end, we resort to deep knowledge distillation (KD)~\cite{nips15-kd}, initially developed for transferring ``dark'' knowledge in a cumbersome deep model to a smaller model. We exploit the KD technique with a novel motivation of simultaneously transferring knowledge from the source and domain-adapted models into another model to make that specific model effective for the expanded domain.

In sum, our contributions are as follows:
\begin{itemize}
	\item We propose UDE as a new task. By simultaneously considering the performance on the source and target domains, UDE is more practical yet more challenging than UDA. The new task allows us to directly measure to what extent a model is domain-invariant. Moreover, to the best of our knowledge, this paper is the first to systematically document the performance of the current UDA models on the source domain, empirically revealing that they are essentially domain-\emph{specific} rather than domain-invariant. 
    \item We propose \textit{Knowledge Distillation Domain Expansion} (KDDE), a general method for the UDE task. Its domain-adaptation module can be instantiated with any existing model. Moreover, our knowledge distillation based learning mechanism allows KDDE to optimize a single objective wherein the source and target domains are equally treated. Note that the knowledge distillation technique used by this paper is not new by itself. We adopt it as a proof-of-concept solution for UDE, developed based on a teacher-student framework. While built upon existing components, KDDE provides a principled approach to leveraging previous UDA models for domain expansion, even when multi-domain shifts exist.

    \item Extensive experiments on two major benchmarks, \ie Office-Home~\cite{officehome} and DomainNet~\cite{domainnet}, show that KDDE outperforms four competitive baselines, \ie DDC, DANN, DAAN, and CDAN, for the UDA and UDE tasks both. Besides, experiments on cross-device OCT image classification show a high potential of the proposed method for improving the generalization ability of a medical image classification system in a real-world multi-device scenario.
\end{itemize}

\section{Related Work} \label{sec:related}

As we have noted in Section \ref{sec:intro}, prior work on UDE does not exist. Nonetheless, UDE relies on UDA techniques~\cite{ddc, jmlr16-dann, nips18-cdan, mm19-jada, icml17-jan, eccvw16-deep-coral}. The proposed KDDE model also benefits from progress in knowledge distillation based deep transfer learning~\cite{nips15-kd,iclr2015-fitnets,zhang2018deep_mutual,mm19-sp-kd}. Therefore, we review briefly recent developments regarding the two topics.

\subsection{Deep Unsupervised Domain Adaptation}

Depending on how domain discrepancy is modeled, we categorize deep learning methods for UDA into two categories, \ie metric-based methods~\cite{ddc,icml15-dan,icml17-jan,eccvw16-deep-coral,domainnet,cvpr19-can} and adversarial methods~\cite{jmlr16-dann,tzeng2017adversarial,icdm19-daan,nips18-cdan,mm19-jada}. As a representative work of the first category, DDC~\cite{ddc} measures domain discrepancy in terms of the Euclidean distance between mean feature vectors of the source and target domains. By jointly minimizing the classification loss on the labeled source domain and the inter-domain distance, DDC aims to learn discriminative yet domain-invariant feature representations. While DDC considers only the last feature layer of the underlying classification network, Joint Adaptation Network (JAN)~\cite{icml17-jan} measures domain discrepancy on multiple task-specific layers. Deep CORAL~\cite{eccvw16-deep-coral} utilizes second-order statistics, learning to reduce the divergence between the covariance matrices of the two domains. Although the above models are end-to-end, their metrics for domain discrepancy have to be empirically predefined, which could be suboptimal.


To bypass the difficulty in specifying a proper discrepancy metric, several methods that resort to adversarial learning have been developed~\cite{jmlr16-dann,nips18-cdan,cvpr2018-mcd,icdm19-daan,mm19-jada}. 
Domain Adversarial Neural Network (DANN)~\cite{jmlr16-dann} introduces a domain classifier as a discriminator, while its feature extractor tries to generate domain-invariant features to confuse the discriminator.
Dynamic Adversarial Adaptation (DAAN)~\cite{icdm19-daan} improves DANN by introducing multiple concept-specific discriminators to dynamically weigh the importance of marginal and conditional distributions. 
In order to align both learned features and predicted classes, Conditional Discriminative Adversarial Network (CDAN)~\cite{nips18-cdan} extends DANN by taking multilinear conditioning of feature representations and classification results as the input of its discriminator.
Different from CDAN that uses a feed-forward network as its discriminator, Maximum Classifier Discrepancy (MCD)~\cite{cvpr2018-mcd} builds a CNN network with two classification branches, and exploits the discrepancy between their output to determine whether a given example is from the source or target domain. Joint Adversarial Domain Adaptation (JADA)~\cite{mm19-jada} extends MCD by adding an additional domain classifier to achieve class-wise and domain-wise alignments. 
While all the above works concentrate on the target domain, our work provides a generic approach to improving their performance on the expanded domain.

%


\subsection{Deep Knowledge Distillation}
Deep knowledge distillation is originally proposed to transfer ``dark'' knowledge in a cumbersome deep model or an ensemble to a smaller model~\cite{nips15-kd}. 
Compared with the big model, the small model trained by its own typically has a lower accuracy. Knowledge distillation provides a principled mechanism to let the small model learn as a student from the big model as a teacher. In particular, the student mimics the teacher's behavior by minimizing the Kullback-Leibler divergence or the cross-entropy loss between the output of the two models~\cite{nips15-kd,zhang2018deep_mutual}. For its general applicability, knowledge distillation has been widely used in varied tasks such as object detection~\cite{nips17-kd-od}, pose regression~\cite{iccv19-pose-kd}, semantic segmentation~\cite{cvpr19-kd-seg}, and saliency prediction~\cite{mm19-sp-kd}.
Not surprisingly, the technology has been investigated for domain adaptation in the context of speech recognition~\cite{asami2017domain,meng2018adversarial,meng2019domain} and image recongition~\cite{self-ensembling,iccv19-self-training}. As we target at domain expansion, we leverage knowledge distillation in a different manner, both conceptually and technically.

\section{Our Method} \label{sec:method}

\subsection{Problem Formalization}

We consider multi-class visual categorization, where a specific example $x$ belongs to one of $k$ predefined visual concepts. Ground truth of $x$ is indicated by $y$, a $k$-dimensional one-hot vector. A deep visual classification network $G$ classifies $x$ by first employing a feature extractor $F$ to obtain a vectorized feature representation $z$ from its raw pixels. Then, $z$ is fed into a $k$-way classifier $C$ to produce a categorical probability vector $\hat{y}$, where the value of its $i$-th dimension is the probability of the example belonging to the $i$-th concept, \ie
\begin{equation} \label{eq:clf}
\left\{
\begin{array}{lll}
z & = & F(x), \\
\hat{y} & = & C(z).
\end{array} \right.
\end{equation}
 Such a paradigm as expressed in Eq. \ref{eq:clf} remains valid to this day, even though $F$ and $C$ have now been jointly deployed and end-to-end trained by deep learning.
%

As UDE is derived from UDA, we adopt common notations from the latter for the ease of consistent description. 
For both UDE and UDA, we have access to a set of $n_s$ labeled training examples $\{(x_{s,i}, y_{s,i})\}_{i=1}^{n_s}$ from a source domain $D_s$ and a set of $n_t$ unlabeled training examples $\{x_{t,i}\}_{i=1}^{n_t}$ from a target domain $D_t$. However, different from UDA that focuses on $D_t$, UDE treats the expanded domain $D_s + D_t$ as its ``target'' domain. Therefore, our goal is to train a unified model that can accurately classify novel examples regardless of their original domains.



\subsection{UDE by Knowledge Distillation} \label{ssec:kdde_model}

\begin{figure*}[tbh!]
\setlength{\abovecaptionskip}{5pt}
\centering
\includegraphics[width=\textwidth]{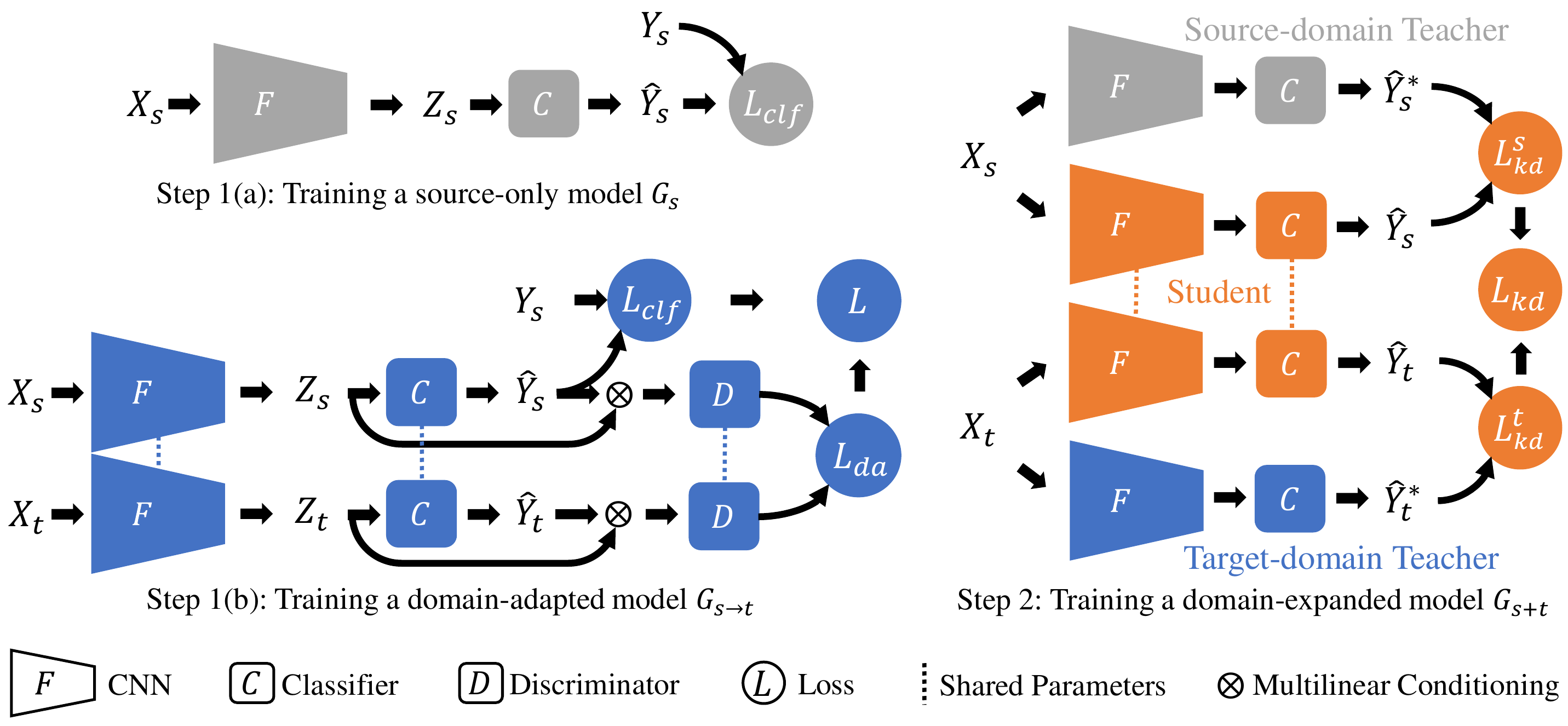}
\caption{\textbf{Conceptual diagram of the proposed KDDE method for UDE}. KDDE is run in two steps. First, a model $G_s$ is trained on the \textit{labeled} source domain by standard supervised learning, while another model $G_{s\rightarrow t}$ for the \textit{unlabeled} target domain is trained by unsupervised domain adaptation (here CDAN~\cite{nips18-cdan} as a running example). Then, knowledge distillation is performed to transfer ``dark'' knowledge from the two domain-specific models into a student model $G_{s+t}$ to make it applicable to both domains.}
\label{fig:method}
\end{figure*}

We propose a knowledge distillation based method for UDE, which we term \textit{KDDE}. As illustrated in Fig. \ref{fig:method}, KDDE is performed in two steps. In the first step, two domain-specific classifiers, denoted as $G_s$ and $G_{s\rightarrow t}$, are trained for the source and target domains, respectively. Note that we use the notation $s\rightarrow t$ to emphasize that as $D_t$ is unlabeled, any domain-adapted classifier shall departure from $D_s$. In the second step, we treat $G_s$ and $G_{s\rightarrow t}$ as two teacher models and transfer their ``dark'' knowledge into a student model $G_{s+t}$ via a knowledge distillation process. In particular, by mimicking $G_s$ for classifying examples from $D_s$ and $G_{s\rightarrow t}$ for $D_t$, $G_{s+t}$ essentially becomes domain-invariant.  We detail the two steps as follows.

\subsubsection{Step 1(a): Training a source-only model $G_s$}

Learning a classifier from labeled data is relatively simple. We adopt the cross-entropy loss, a common classification loss used for training a multi-class deep neural network. Given $(X_s,Y_s)$ as $\{(x_{s,i}, y_{s,i})\}_{i=1}^{n_s}$, the classification loss is written as
\begin{equation} \label{eq:clf-loss}
L_{clf}(X_s, Y_s, G) = \frac{1}{n_s} \sum_{(x_s,y_s)\in (X_s,Y_s)} y_s \log (G(x_s)).
\end{equation} 
By minimizing $L_{clf}(X_s, Y_s, G)$, we obtain a domain-specific model for $D_s$ as
\begin{equation} \label{eq:g_s}
G_s = \argmin{G}  L_{clf}(X_s, Y_s, G).
\end{equation}


\subsubsection{Step 1(b): Training a domain-adapted model $G_{s\rightarrow t}$}



Recall that KDDE, as a two-stage solution, is agnostic to the implementation of a specific UDA model used in its first stage.
Hence, for obtaining $G_{s\rightarrow t}$, any method for unsupervised domain adaptation can, in principle, be adopted. 
A typical process of model adaptation tries to strike a proper balance between a model's discriminability, as reflected by the classification loss on $D_s$, and its domain-invariant representation ability, as measured by inter-domain discrepancy between $D_s$ and $D_t$. More formally, we have 
\begin{equation}\label{eq:general-da}
G_{s\rightarrow t} = \argmin{G} \underbrace{L_{clf}(X_s, Y_s, G)}_{\mbox{discriminative}} + \lambda \cdot \underbrace{L_{da}(X_s, X_t, G)}_{\mbox{domain-invariant}},
\end{equation}
where $L_{da}$ is a domain discrepancy based loss, and $\lambda$ is a positive hyper-parameter. Algorithm \ref{alg:da} describes the domain adaptation process at a high level. In what follows, we use the state-of-the-art CDAN model~\cite{nips18-cdan} as a running example to instantiate $L_{da}$. 

\begin{algorithm}
\caption{Training a domain-adapted model $G_{s\rightarrow t}$}
\label{alg:da}
\KwIn{$(X_{s}, Y_{s})$, $X_{t}$}
\KwOut{$G_{s\rightarrow t}$}
Set hyper-parameters: $\lambda, MAX\_EPOCHS, BATCH\_SIZE$\;
Compute the number of iterations per epoch $MAX\_ITERS$, given $|X_s|$ and  $BATCH\_SIZE$\;
Initialize {$G_{s\rightarrow t}$} with an ImageNet-pretrained model\;
\For{$i\leftarrow 1,\ldots, MAX\_EPOCHS$}{
    \For{$j\leftarrow 1, \ldots, MAX\_ITERS$}{
      Sample a mini-batch $(X_{s,j}, Y_{s,j})$ from $(X_{s}, Y_{s})$\;
      Sample a mini-batch $X_{t,j}$ from $X_t$\;
      Compute the classification loss $L_{clf}(X_{s,j}, Y_{s,j}, G_{s\rightarrow t})$ \;
      Compute the inter-domain discrepancy $L_{da}(X_{s,j}, X_{t,j}, G_{s\rightarrow t})$ \;
      Update $G_{s\rightarrow t}$ by back propagation based on $L_{clf}+\lambda*L_{da}$
      }
    }
\end{algorithm}



CDAN reduces domain discrepancy by adversarial learning, where a feed-forward neural network is used as a discriminator $D$ to disentangle the source examples $X_s$ from the target examples $X_t$. Different from previous adversarial learning based methods where only the intermediate features $F(x_s)$ and $F(x_t)$ are considered, CDAN uses multilinear conditioning of the features and class prediction $F(x)\otimes \hat{y}$ as the input of $D$. The output of $D$ is the probability of a given example coming from the source domain. Accordingly, we have the discriminator reward $R$ as 
\begin{equation}
R(X_s,X_t,G,D) = \sum_{x_s \in X_s} \log D(F(x_s)\otimes \hat{y}_s) + \sum_{x_t \in X_t} \log (1 - D(F(x_t) \otimes \hat{y}_t))
\end{equation} 
The training process of CDAN is implemented as a two-player minimax game as follows:
\begin{equation} \label{eq:cdan}
    (G_{s\rightarrow t}, D)=\argmin{G} \underset{D}{\max}\; L_{clf}(X_s, Y_s, G) + \lambda \cdot R(X_s, X_t, G, D).
\end{equation}

\subsubsection{Step 2. Training a domain-expanded model $G_{s+t}$}

Given the domain-specific models $G_s$ and $G_{s\rightarrow t}$, we now transfer their capabilities in their own domains into a new model $G_{s+t}$ by knowledge distillation. 

In a standard scenario where one wants to distill the knowledge in a big teacher model into a relatively small student model~\cite{nips15-kd}, both ground-truth hard labels and soft labels predicted by the teacher model are available for computing the distillation loss. By contrast, in the setting of UDE, the expanded domain has only partial ground-truth labels by definition. More importantly, in order to make $G_{s+t}$ domain-invariant, we shall not only treat $G_s$ and $G_{s\rightarrow t}$ equally, but also exploit training examples from $D_s$ and $D_{s\rightarrow t}$ in the same manner. To that end, we opt to compute the distillation loss fully based on the soft labels. 

Specifically, we adopt the Kullback-Leibler (KL) divergence, as previously used to quantify how a student's output matches with its teacher~\cite{nips15-kd,zhang2018deep_mutual}. Per training example, the soft labels produced by the teacher / student model is essentially a probability distribution over the $k$ concepts. The KL divergence provides a natural measure of how the probability distribution produced by the student is different from that of the teacher, making it a popular loss for knowledge distillation \cite{nips15-kd,zhang2018deep_mutual,mirzadeh2020improved,tian2019contrastive,goldblum2020adversarially,zhang2019your,shi2019knowledge}. Indeed, our ablation study in Section \ref{sssec:kd-loss} shows that the KL divergence loss is better than other losses such as cross-entroy and $l_2$. Let $P(G(X))$ be a $k$-dimensional categorical distribution estimated based on soft labels of an example set $X$ produced by a specific network $G$. Accordingly, the KL divergence from $G$ to each of the two teachers is defined as $KL(P(G_s(X_s)) || P(G(X_s)))$ and $KL(P(G_{s\rightarrow t}(X_t)) || P(G(X_t)))$, respectively. 
The knowledge distillation loss $L_{kd}$ is defined as 
\begin{equation} \label{eq:loss-kd}
L_{kd}(X_s, X_t, G_s, G_{s\rightarrow t}, G) = KL(P(G_s(X_s)) || P(G(X_s)))  + KL(P(G_{s\rightarrow t}(X_t)) || P(G(X_t))).
\end{equation}

Note that the $KL$ terms in Eq. \ref{eq:loss-kd} are practically computed by a mini-batch approach. As demonstrated in Fig. \ref{fig:method}, in each iteration two mini-batches are independently and randomly sampled from $D_s$ and $D_t$. They are then fed into $G_s$ and $G_{s\rightarrow t}$ to get the soft labels, which are used to approximate $P(G_s(X_s))$ and $P(G_{s\rightarrow t}(X_t))$, respectively. Meanwhile, the two batches are also fed to the student model. Minimizing $L_{kd}$ lets the student model mimic the teachers' behaviors on both $D_s$ and $D_t$. 
Consequently, we obtain the domain-invariant model as 
\begin{equation} \label{eq:general-de}
    G_{s+t}=\argmin{G} L_{kd}(X_s, X_t, G_s, G_{s\rightarrow t}, G).
\end{equation}

A high-level description of the training process is given in Algorithm \ref{alg:kdde}. Note that during training, we need to store three models (two teachers $G_s$ and $G_{s\rightarrow t}$ and one student $G_{s+t}$). We consider such storage overhead affordable as one typically has access to more computational resources in the training stage than in the inference stage. Moreover, as the teachers are only used to product the soft labels, their weights are frozen, meaning the GPU footprint is much less than simultaneously training all the three models. Also note that in the inference stage. $G_{s+t}$ has the same model size and computation overhead as its teachers. Hence, the proposed method is feasible in real-world scenarios.

\begin{algorithm}
\caption{Training a domain-expanded model $G_{s+t}$ by KDDE}
\label{alg:kdde}
\KwIn{$X_s, X_t, G_s, G_{s\rightarrow t}$}
\KwOut{$G_{s+t}$}
Set hyper-parameters: $MAX\_EPOCHS, BATCH\_SIZE$\;
Compute $MAX\_ITERS$ given $max(|X_s|,|X_t|)$ and  $BATCH\_SIZE$\;
Initialize {$G_{s+t}$} with an ImageNet-pretrained model\;
\For{$i\leftarrow 1,MAX\_EPOCH$}{
    \For{$j\leftarrow 1,MAX\_ITER$}{
        Sample $X_{s,j}$ from $X_s$\; 
        Sample $X_{t,j}$ from $X_t$\;
        Compute $L_{kd}(X_{s,j}, X_{t,j}, G_s, G_{s\rightarrow t}, G_{s+t})$ using Eq. \ref{eq:loss-kd}\;
        Update $G_{s+t}$ by back propagation based on $L_{kd}$\;
        }
    }
\end{algorithm}

\subsection{Theoretical Analysis}

Comparing Eq. \ref{eq:general-da} and Eq. \ref{eq:general-de}, we see that UDA essentially tries to simultaneously optimize two distinct and sometimes conflictive objectives, \ie discriminability and domain-invariance. By contrast, our KDDE optimizes a single objective. We believe such a property improves the cross-domain generalization ability of KDDE. Nonetheless, because $G_{s+t}$ is learned from the source model $G_s$ and the domain-adapted model $G_{s\rightarrow t}$ in an unbiased manner, domain expansion is obtained at the cost of certain performance drop in the source domain. In other words, $G_{s+t}$ will be less effective than its teacher $G_{s}$ on the source domain, but perform better on the expanded domain, which is the goal of this research.

As for the target domain, $G_{s+t}$ will be better than its teacher $G_{s\rightarrow t}$. According to Ben-David \etal \cite{ben2007analysis}, the target domain error of a UDA model is bounded mainly by its classification error in the source domain and the divergence between the induced source marginal and the induced target marginal. In theory, a UDA model shall be trained to simultaneously minimize the two terms and reduce accordingly the up bound of the target domain error. In practice, however, the classification error in the source domain is not effectively reduced when compared to the source-only model. This is confirmed by our experiments in Section \ref{sec:exp} that a number of present-day UDA models including DDC~\cite{ddc}, DANN~\cite{jmlr16-dann}, DAAN~\cite{icdm19-daan} and CDAN~\cite{nips18-cdan} suffer performance loss in the source domain. With knowledge distillation, $G_{s+t}$ effectively integrates the merits of $G_{s\rightarrow t}$ for minimizing the domain divergence and $G_s$ for minimizing the source error, and thus lowers the up bound of the target domain error. 

\begin{figure*}[tbh!]
\setlength{\abovecaptionskip}{5pt}
\centering
\includegraphics[width=0.55\textwidth]{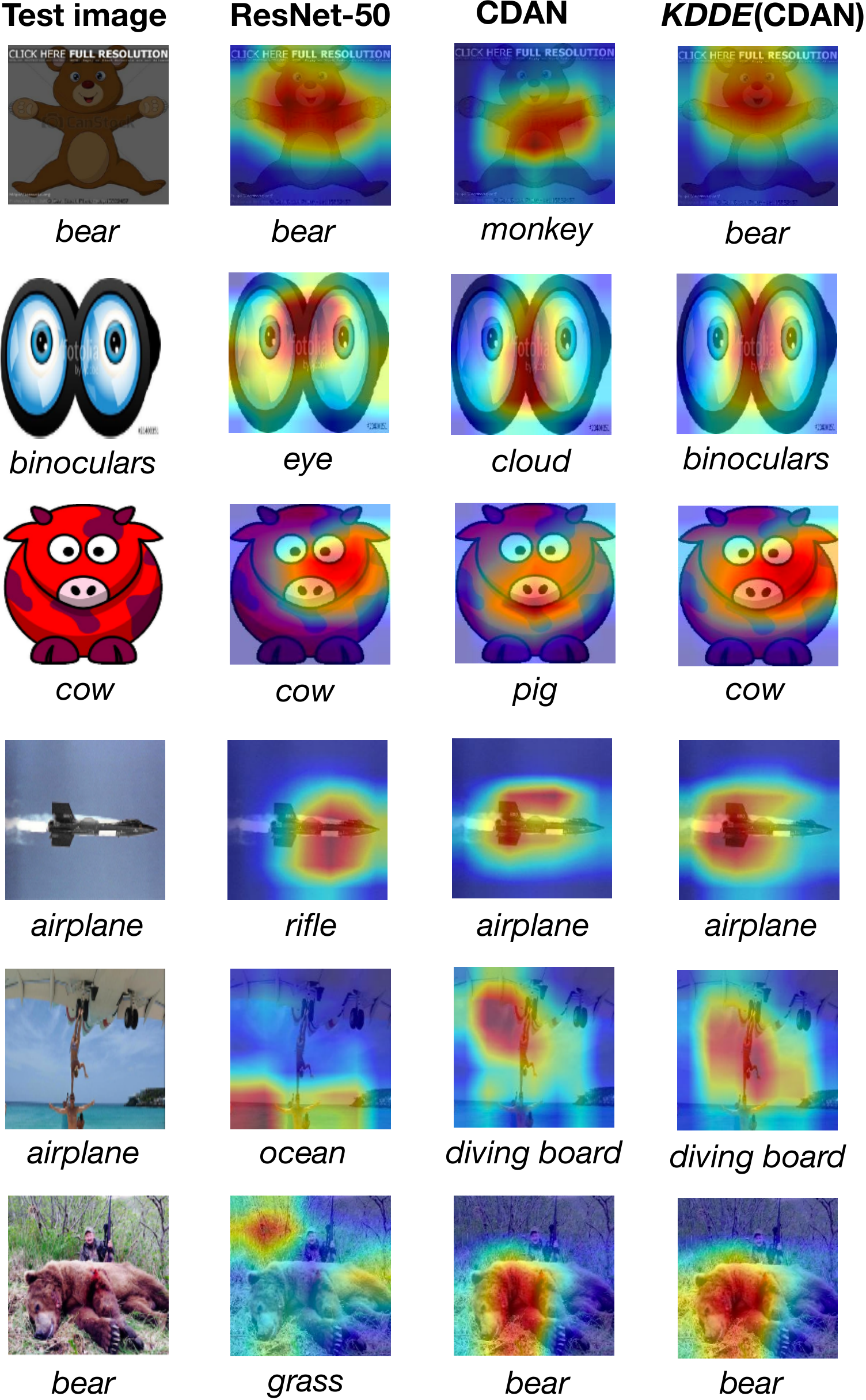}
\caption{\textbf{Grad-CAM heatmaps of ResNet-50, CDAN and KDDE}. Heatmaps highlight regions important for a model to make predictions. Test images in the first three rows are from the source domain (\textit{clipart}), while images in the last three rows are from the target domain (\textit{real}). Activated regions of KDDE ($G_{s+t}$) match well with those of ResNet-50 ($G_s$) on the source domain and those of CDAN ($G_{s\rightarrow t}$) on the target domain, suggesting the ability of KDDE to adaptively learn from the two domain-specific models. Data from DomainNet.}
\label{fig:heatmaps}
\end{figure*}

Through knowledge distillation, KDDE injects the dark knowledge of the source-only model $G_s$, which performs well on the source domain, and the domain-adapted model $G_{s\rightarrow t}$, which is supposed to perform well on the target domain, into a single model $G_{s+t}$. The effect of knowledge distillation on $G_{s+t}$ is visualized in Fig. \ref{fig:heatmaps}. By contrast, although CDAN also uses one classifier for both domains, the classifier is essentially $G_{s\rightarrow t}$ used in KDDE. Hence, it is less effective than $G_{s+t}$ to handle the expanded domain. Also notice that the purpose of knowledge distillation is not to reduce the difference between the two teacher models, see Eq. \ref{eq:loss-kd}. Therefore, KDDE is conceptually different from Maximum classifier discrepancy (MCD) \cite{cvpr2018-mcd}, which is to reduce the discrepancy between two domain-adapted classifiers via a novel adversarial learning mechanism.

\section{Experiments} \label{sec:exp}



\subsection{Datasets} \label{ssec:exp-data}

We evaluate the effectiveness of the proposed KDDE model on two public benchmarks, \ie Office-Home~\cite{officehome} and DomainNet~\cite{domainnet}, and a private dataset OCT-11k.

\subsubsection{Office-Home}

The Office-Home dataset contains 15,588 images of 65 object categories typically found in office and home settings, \eg chair, table, and TV. Images were collected from the following four distinct domains, \ie \textit{Art} (A), \textit{Clipart} (C), \textit{Product} (P) and \textit{Real world} (R). As the dataset is previously used in the domain adaption setting~\cite{nips18-cdan, wang2019transnorm, pinheiro2018unsupervised, wang2019transatten} that considers only performance on the target domain, no test set is provided for the source domain. So to evaluate UDE, for each domain, we randomly divide its images into two disjoint subsets, one for training and the other for test, at a ratio of 1:1, see Table \ref{tab:data}. An expanded domain of C+P means using Clipart as the (labeled) source domain, which is expanded with Product as the (unlabeled) target domain. Due to such asymmetric nature of the label information, P+C differs from C+P. Pairing the individual domains results in 12 distinct UDE tasks in total.

\begin{table}[tbh!]
\renewcommand\arraystretch{1.2}
\centering
\caption{\textbf{Three datasets used in our experiments for UDA and UDE}. Note that different from the traditional setting of UDA which uses all examples in the source domain for training, the setting of UDE divides the source examples into two disjoint parts, \textit{training} and \textit{test}. This allows us to evaluate both UDA and UDE models on the original source domain, which is fully ignored by the literature of UDA.}
\label{tab:data}
\scalebox{0.8}{
\begin{tabular}{@{}|l|l|r|r|r|@{}}
\hline
 \multirow{2}{*}{\textbf{Dataset}} & \multirow{2}{*}{\textbf{Domains}} & \multicolumn{3}{c|}{\textbf{Images}} \\
 \cline{3-5}
 & & \emph{total} & \emph{training} & \emph{test} \\
 \hline
\multirow{4}{*}{\specialcell{Office-Home~\cite{officehome} \\ 15,588 images \\ 65 classes \\ }} & A: Art & 2,427  & 1,201 & 1,226  \\
\cline{2-5}
& C: Clipart & 4,365 & 2,165 & 2,200   \\
\cline{2-5}
& P: Product & 4,439 & 2,201 & 2,238  \\
\cline{2-5}
& R: Real\_world & 4,357  & 2,161& 2,196 \\
\hline
\multirow{4}{*}{\specialcell{Subset of \\ DomainNet~\cite{domainnet} \\ 362,470 images \\ 345 classes \\ }} & c: clipart & 48,129 & 33,525 & 14,604 \\
\cline{2-5}
& p: painting & 72,266 & 50,416 & 21,850\\
\cline{2-5}
& r: real & 172,947 & 120,906 & 52,041\\
\cline{2-5}
& s: sketch & 69,128 &  48,212 & 20,916 \\
\hline
\multirow{2}{*}{\specialcell{OCT-11k (our private dataset) \\ 11,800 images, 2 classes}} 
& Z: Zeiss & 5,900 & 5,000 & 900 \\
\cline{2-5}
& T: Topcon & 5,900 & 5,000 & 900\\
\hline
\end{tabular}
}
\end{table}

\subsubsection{DomainNet}

DomainNet is a recent large-scale benchmark dataset used in  the Visual Domain Adaptation Challenge at ICCV 2019\footnote{\url{https://ai.bu.edu/visda-2019}}. Compared with Office-Home, DomainNet contains a much larger number of 345 object categories with more intra-class diversity and inter-class ambiguity in visual appearance. As such, it is difficult to obtain high classification accuracy even within a narrow domain. The full set of DomainNet has six domains, \ie, \textit{clipart} (c), \textit{infograph} (i), \textit{painting} (p), \textit{quickdraw} (q),  \textit{real} (r) and \textit{sketch} (s). Saito \etal~\cite{domainnet-sub} exclude inforgraph and quickdraw from their study as they find annotations of these two domains are over noisy. We follow their setup, experimenting with the four other domains. 

\subsubsection{OCT-11k} \label{ssec:dataset-oct}

In order to evaluate the effectiveness of our proposed method for medical image classification in a cross-device setting, we build a set of 11,800 OCT B-scan images, half of which was collected by \textit{Zeiss Cirrus OCT} and the other half from \textit{Topcon 2000FA OCT}. 
All the \textit{Zeiss} images and a subset of 900 \textit{Topcon} images were labeled by experts into two classes, namely positive and negative. An image was labeled as positive if certain pathological anomaly such as \textit{macular edema}, \textit{macular hole}, \textit{drusen}, and \textit{choroidal atrophy} is present. Accordingly, we treat \textit{Zeiss} as the source domain and \textit{Topcon} as the target domain. To let the expanded domain contain an equal number of test examples from the individual domains, we randomly sample a subset of 900 \textit{Zeiss} images as the test set of the source domain. The \textit{Zeiss} test set contains 581 positives and 319 negatives, while the \textit{Topcon} test set contains 471 positives and 429 negatives. 

\subsection{Implementation} \label{ssec:implement}

\subsubsection{Baselines}

We compare with the following state-of-the-art domain adaptation models.
\begin{itemize}
	\item DDC~\cite{ddc}: A classical deep domain adaptation model that minimizes domain discrepancy measured in light of first-order statistics of the deep features.
	\item DANN~\cite{jmlr16-dann}: Among the first to obtain domain-invariant deep features by adversarial learning. 
	\item DAAN~\cite{icdm19-daan}: An adversarial learning based domain adaptation model, with discriminator networks per concept. For its high demand for GPU memory, we are unable to run DAAN on DomainNet. 
	\item CDAN~\cite{nips18-cdan}: Also based adversarial learning, using multilinear conditioning of deep features and classification results as the input of its discriminator.
\end{itemize}

Recall that DDC and CDAN are representatives for metric-based and adversarial methods, respectively. Hence, we instantiate $G_{s\rightarrow t}$ using the two models separately, resulting in two variants of KDDE, denoted as \emph{KDDE}(DDC) and \emph{KDDE}(CDAN). 

We implement the source model $G_s$ using ResNet-50 that is trained exclusively on the source domain. A model for UDA / UDE shall outperform this baseline on the target / expanded domain. For fair comparison between distinct models, we also use ResNet-50 as their backbones. 



\subsubsection{Model training}

We run all experiments with PyTorch~\cite{paszke2019pytorch}. We start with ResNet-50 pre-trained on ImageNet. SGD is used for training, with momentum of $0.9$ and weight decay of $0.0005$. The initial learning rate of ResNet-50, DDC, DANN, DAAN and CDAN is empirically set to $0.001$, and $0.005$ for KDDE. For the adversarial methods, \ie DANN, DAAN and CDAN, we adopt the inverse-decay learning rate strategy from~\cite{nips18-cdan}. As for ResNet-50, DDC and KDDE, the learning rate is decayed by $0.1$ every $30$ epochs on Office-Home and every $10$ epochs on DomainNet, as the latter has much more training examples and thus more iterations per epoch. For the same reason, for each model we train 100 epochs on Office-Home and a less number of 30 epochs on DomainNet. 

For DDC, DANN and DAAN, the trade-off parameter $\lambda$ is empirically set to be $10$, $1$, and $1$, respectively. As for CDAN, we follow the original paper~\cite{nips18-cdan} to adjust $\lambda$ dynamically.



\subsubsection{Evaluation protocol}

For overall performance, we report accuracy, \ie the percentage of test images correctly classified, as commonly used for evaluating multi-class image classification. For an expanded domain, \eg A+C, its test set is the union of the test sets of the Art and Clipart domains. To cancel out data imbalance, the accuracy of the expanded domain is obtained by averaging over the two individual domains. The performance of a specific concept is measured by F1-score, the harmonic mean of precision and recall. For binary classification on OCT-11k, we additionally report the area under the ROC curve (AUC).





\subsection{Results on Office-Home}

Table \ref{tab:exp_total} reports the overall performance, with per-task results detailed in Table \ref{tab:office_part1} and Table \ref{tab:office_part2}. Note that although the performance gap appears to be relatively small, see also the performance reported in \cite{icml17-jan,icdm19-daan}, the significance of the gap shall not be underestimated due to the challenging nature of the UDA / UDE tasks.

\begin{table*}[tbh!]
	\centering
	\caption{\textbf{Overall performance of distinct models on Office-Home and DomainNet}. Performance metric is accuracy, shown in percentages. We use ResNet-50 as a reference. Performance increases (decreases) over this reference are shown in \textcolor{red}{red} (\textcolor{darkgreen}{green}). Note that training DAAN on DomainNet is beyond our computational capacity, so its performance on DomainNet is unavailable. Better performance of KDDE against alternatives on the target / expanded domain justifies its effectiveness for UDA / UDE.}
	\label{tab:exp_total}
	\scalebox{0.8}	{
	\begin{tabular}{|l|l|l|l|}
	\hline
	\multirow{2}{*}{\textbf{Model}} & \multicolumn{3}{c|}{\textbf{Office-Home}} \\ \cline{2-4} 
	 & \textit{Source domain} & \textit{Target domain} & \textit{Expanded domain} \\ \hline
	ResNet-50 & \textbf{82.57} & 57.49 & 70.03 \\ \hline
	DANN & 81.42$\downarrow$\textcolor{darkgreen}{-1.15} & 60.89$\uparrow$\textcolor{red}{+3.40} & 71.16$\uparrow$\textcolor{red}{+1.13} \\ \hline
	CDAN & 80.54$\downarrow$\textcolor{darkgreen}{-2.03} & 61.85$\uparrow$\textcolor{red}{+4.36} & 71.20$\uparrow$\textcolor{red}{+1.17} \\ \hline
	DDC & 82.22$\downarrow$\textcolor{darkgreen}{-0.35} & 60.61$\uparrow$\textcolor{red}{+3.12} & 71.41$\uparrow$\textcolor{red}{+1.38} \\ \hline
	DAAN & 82.37$\downarrow$\textcolor{darkgreen}{-0.20} & 60.78$\uparrow$\textcolor{red}{+3.29} & 71.57$\uparrow$\textcolor{red}{+1.54} \\ \hline
	\textit{KDDE}(DDC) & \textbf{82.57}$\uparrow$\textcolor{red}{+0.006} & 61.62$\uparrow$\textcolor{red}{+4.13} & 72.10 $\uparrow$\textcolor{red}{+2.07} \\ \hline
	\textit{KDDE}(CDAN) & 81.44$\downarrow$\textcolor{darkgreen}{-1.13} & \textbf{63.90}$\uparrow$\textcolor{red}{+6.41} & \textbf{72.67}$\uparrow$\textcolor{red}{+2.64} \\ \hline
	\multirow{2}{*}{\textbf{Model}} & \multicolumn{3}{c|}{\textbf{DomainNet}} \\ \cline{2-4} 
	 & \textit{Source domain} & \textit{Target domain} & \textit{Expanded domain} \\ \hline
	ResNet-50 & \textbf{74.59} & 41.49 & 58.04 \\ \hline
	DANN & 69.37$\downarrow$\textcolor{darkgreen}{-5.22} & 44.53$\uparrow$\textcolor{red}{+3.04} & 56.95$\downarrow$\textcolor{darkgreen}{-1.09} \\ \hline
	CDAN & 69.73$\downarrow$\textcolor{darkgreen}{-4.86} & 45.21$\uparrow$\textcolor{red}{+3.72} & 57.47$\downarrow$\textcolor{darkgreen}{-0.57} \\ \hline
	DDC & 72.44$\downarrow$\textcolor{darkgreen}{-2.15} & 46.20$\uparrow$\textcolor{red}{+4.71} & 59.32$\uparrow$\textcolor{red}{+1.28} \\ \hline
	DAAN & N.A. & N.A. & N.A. \\ \hline
	\textit{KDDE}(DDC) & 73.78$\downarrow$\textcolor{darkgreen}{-0.81} & \textbf{48.04}$\uparrow$\textcolor{red}{+6.55} & \textbf{60.91}$\uparrow$\textcolor{red}{+2.87} \\ \hline
	\textit{KDDE}(CDAN) & 72.98$\downarrow$\textcolor{darkgreen}{-1.61} & 47.65$\uparrow$\textcolor{red}{+6.16} & 60.31$\uparrow$\textcolor{red}{+2.27} \\ \hline
	\end{tabular}
	}
\end{table*}

\begin{table*}[tbh!]
	\caption{\textbf{Performance on Office-Home}. Art (A) and Clipart (C) are used as the source domain, respectively. The notation $A \rightarrow C$ means using Clipart (C) as the target domain and consequently resulting in an expanded domain of A + C. For five out of the six UDA / UDE tasks, KDDE(CDAN) performs the best. }
	\centering
	\label{tab:office_part1}
	\scalebox{0.8}
	{
	\begin{tabular}{|l|c|c|c|c|c|c|c|c|c|}
	\hline
	\multirow{2}{*}{\textbf{Model}} & \multicolumn{3}{c|}{A$\rightarrow$C} & \multicolumn{3}{c|}{A$\rightarrow$P} & \multicolumn{3}{c|}{A$\rightarrow$R} \\ \cline{2-10} 
	 & A & C & A+C & A & P & A+P & A & R & A+R \\ \hline
	ResNet-50 & \textbf{75.20} & 45.23 & 60.22 & \textbf{75.20} & 58.45 & 66.83 & \textbf{75.20} & 69.35 & 72.28 \\ \hline
	DDC & 72.51 & 49.09 & 60.80 & 73.65 & 62.60 & 68.13 & 73.90 & 70.86 & 72.38 \\ \hline
	DANN & 71.29 & \textbf{49.23} & 60.26 & 73.33 & 60.99 & 67.16 & 74.31 & 70.17 & 72.24 \\ \hline
	DAAN & 73.98 & 48.95 & \textbf{61.47} & 73.98 & 64.30 & 69.14 & 74.39 & 71.17 & 72.78 \\ \hline
	CDAN & 70.96 & 46.73 & 58.85 & 71.78 & 64.61 & 68.20 & 72.59 & 70.63 & 71.61 \\ \hline
	KDDE(DDC) & 73.08 & 48.86 & 60.97 & 74.39 & 63.67 & 69.03 & 74.96 & 71.22 & 73.09 \\ \hline
	KDDE(CDAN) & 70.07 & 48.77 & 59.42 & 72.19 & \textbf{66.71} & \textbf{69.45} & 73.98 & \textbf{72.40} & \textbf{73.19} \\ \hline
	\multirow{2}{*}{\textbf{Model}} & \multicolumn{3}{c|}{C$\rightarrow$A} & \multicolumn{3}{c|}{C$\rightarrow$P} & \multicolumn{3}{c|}{C$\rightarrow$R} \\ \cline{2-10} 
	 & C & A & C+A & C & P & C+P & C & R & C+R \\ \hline
	ResNet-50 & 78.91 & 47.06 & 62.99 & 78.91 & 57.55 & 68.23 & 78.91 & 59.65 & 69.28 \\ \hline
	DDC & \textbf{80.23} & 50.90 & 65.57 & 79.59 & 62.60 & 71.10 & 78.77 & 63.57 & 71.17 \\ \hline
	DANN & 78.27 & 54.08 & 66.18 & 78.86 & 61.71 & 70.29 & 78.91 & 63.02 & 70.97 \\ \hline
	DAAN & 78.64 & 53.34 & 65.99 & 79.86 & 62.29 & 71.08 & 79.82 & 64.12 & 71.97 \\ \hline
	CDAN & 77.82 & 53.34 & 65.58 & 78.00 & 66.13 & 72.07 & 79.09 & 63.93 & 71.51 \\ \hline
	KDDE(DDC) & 80.05 & 54.57 & 67.31 & \textbf{80.68} & 64.97 & 72.83 & 80.18 & 65.16 & 72.67 \\ \hline
	KDDE(CDAN) & 78.27 & \textbf{57.75} & \textbf{68.01} & 79.77 & \textbf{68.68} & \textbf{74.23} & \textbf{80.36} & \textbf{66.67} & \textbf{73.52} \\ \hline
	\end{tabular}
	}
\end{table*}

\begin{table*}[tbh!]
	\caption{\textbf{Performance on Office-Home}. Product (P) and Real\_world (R) are used as the source domain, respectively. For all the six UDA tasks and five out of the fix UDE tasks, KDDE(CDAN) performs the best.}
	\centering
	\label{tab:office_part2}
	\scalebox{0.8}
	{
	\begin{tabular}{|l|c|c|c|c|c|c|c|c|c|}
	\hline
	\multirow{2}{*}{\textbf{Model}} & \multicolumn{3}{c|}{P$\rightarrow$A} & \multicolumn{3}{c|}{P$\rightarrow$C} & \multicolumn{3}{c|}{P$\rightarrow$R} \\ \cline{2-10} 
	 & P & A & P+A & P & C & P+C & P & R & P+R \\ \hline
	ResNet-50 & 92.05 & 50.33 & 71.19 & \textbf{92.05} & 43.86 & 67.96 & 92.05 & 70.31 & 81.18 \\ \hline
	DDC & \textbf{92.09} & 53.67 & 72.88 & 91.55 & 45.05 & 68.30 & 92.49 & 71.95 & 82.22 \\ \hline
	DANN & 90.57 & 53.18 & 71.88 & 89.95 & 47.05 & 68.50 & 91.64 & 72.59 & 82.12 \\ \hline
	DAAN & 91.82 & 53.67 & 72.75 & 91.51 & 44.09 & 67.80 & 92.45 & 72.63 & 82.54 \\ \hline
	CDAN & 91.11 & 53.67 & 72.39 & 88.83 & 49.36 & 69.10 & 91.20 & 73.82 & 82.51 \\ \hline
	KDDE(DDC) & 91.91 & 54.73 & 73.32 & 91.60 & 46.27 & 68.94 & \textbf{92.67} & 73.36 & 83.02 \\ \hline
	KDDE(CDAN) & 91.51 & \textbf{55.55} & \textbf{73.53} & 90.30 & \textbf{49.91} & \textbf{70.11} & 92.45 & \textbf{75.68} & \textbf{84.07} \\ \hline
	\multirow{2}{*}{\textbf{Model}} & \multicolumn{3}{c|}{R$\rightarrow$A} & \multicolumn{3}{c|}{R$\rightarrow$C} & \multicolumn{3}{c|}{R$\rightarrow$P} \\ \cline{2-10} 
	 & R & A & R+A & R & C & R+C & R & P & R+P \\ \hline
	ResNet-50 & 84.11 & 63.62 & 73.87 & \textbf{84.11} & 48.23 & 66.17 & \textbf{84.11} & 76.27 & 80.19 \\ \hline
	DDC & \textbf{84.70} & 64.19 & 74.45 & 82.97 & 52.23 & 67.60 & 83.93 & 77.35 & 80.64 \\ \hline
	DANN & 84.06 & 65.33 & 74.70 & 82.65 & 55.36 & 69.01 & 83.24 & 77.97 & 80.61 \\ \hline
	DAAN & 84.61 & 64.85 & \textbf{74.73} & 83.29 & 52.09 & 67.69 & 84.06 & 77.84 & 80.95 \\ \hline
	CDAN & 82.01 & 64.03 & 73.02 & 80.42 & 55.73 & 68.08 & 82.70 & 80.21 & 81.46 \\ \hline
	KDDE(DDC) & 84.38 & 64.52 & 74.45 & 83.24 & 53.86 & 68.55 & 83.74 & 78.28 & 81.01 \\ \hline
	KDDE(CDAN) & 83.29 & \textbf{65.50} & 74.40 & 81.65 & \textbf{57.73} & \textbf{69.69} & 83.42 & \textbf{81.41} & \textbf{82.42} \\ \hline
	\end{tabular}
	}
\end{table*}

\subsubsection{Performance on the target domain}

All the domain adaption models are found to be better than the source-only ResNet-50. This is consistent with previous works on UDA. Among them, CDAN performs the best, obtaining a relative improvement of 4.36\%. KDDE(CDAN) surpasses CDAN, with accuracy increased from 61.85 to 63.90. KDDE(DDC) is better than DDC, with accuracy increased from 60.61 to 61.62. The results confirm that KDDE is beneficial for the UDA task. 


\subsubsection{Performance on the source domain}

In contrast to their performance on the target domain, the domain adaption models consistently show performance degeneration on the source domain, with their relative loss ranging from 0.20\% (DAAN) to 2.03\% (CDAN). In particular, CDAN as the best domain adaptation model degenerates the most, suggesting that the gain on the target domain is obtained at the cost of affecting the classification ability on the source domain. The use of the proposed KDDE reduces such cost. In particular, KDDE(CDAN) reduces the loss of CDAN from 2.03\% to 1.13\%, while KDDE(DDC) is even comparable to the original ResNet-50 model.




\subsubsection{Performance on the expanded domain}

KDDE(CDAN) performs the best. Moreover, as shown in Table \ref{tab:office_part1} and \ref{tab:office_part2}, for 10 out of all the 12 UDE tasks, KDDE(CDAN) has the highest accuracy, followed by KDDE(DDC). 

To further verify the necessity of KDDE, we compare it with the \emph{Model Selection} method, previously shown in Fig. \ref{fig:conceptual}~(a). Given the two domain-specific models (ResNet-50 and CDAN) trained, model selection classifies a test example using ResNet-50 if the example is deemed to be from the source domain, and using CDAN otherwise. To that end, another ResNet-50 is trained as a domain classifier. Hence, the model selection method requires three ResNet-50 models per task. In addition, we compare with \emph{Model ensemble}, another baseline that combines ResNet-50 and CDAN by late average fusion.  


We select two UDE tasks from Office-Home, namely C$\rightarrow$A and P$\rightarrow$R, considering that the Clipart and Art domains have significant visual difference, while the Product and Real\_world domains look similar. Indeed, this is confirmed by the performance of the domain classifier, which can separate Clipart from Art with an accuracy of 96.56 and a lower accuracy of 85.39 for distinguishing the other two domains. 
As shown in Table \ref{tab:office-two-stage}, KDDE is  better than model selection, which uses three ResNet-50 models. To remove the influence of incorrect domain classification, we also try model selection with ground-truth domain labels, which corresponds to Model selection (Oracle) in Table \ref{tab:office-two-stage}. Again, KDDE is better. Also note that providing ground-truth domain labels does not necessarily lead to better performance. 
Model ensemble has accuracy of 66.78 on C$\rightarrow$A and 83.25 on P$\rightarrow$R. Note that KDDE uses one ResNet50 model while the ensemble requires two ResNet50s. KDDE is better than the ensemble in terms of accuracy, yet uses 50\% less resources at runtime. The results further demonstrate the importance of learning domain-invariant models for UDE.   



\begin{table}[tbh!]
\renewcommand\arraystretch{1}
\centering
\caption{\textbf{KDDE versus Model selection for UDE}. Model selection classifies a test example using ResNet-50 if the example is deemed to be from the source domain, or using CDAN otherwise. } 
\label{tab:office-two-stage}
\scalebox{0.8}{
    \begin{tabular}{|l|c|c|}
    \hline
    \textbf{Model} & C$\rightarrow$A & P$\rightarrow$R \\ \hline
    ResNet-50 & 62.99 & 81.18 \\ \hline
    CDAN & 65.58 & 82.51  \\ \hline
    Model selection (Domain classifier)& 66.29 & 82.71  \\ \hline
    Model selection (Oracle)  & 66.13 & 82.94  \\ \hline
    Model ensemble & 66.78 & 83.25 \\ \hline
    KDDE (CDAN) & \textbf{68.01} & \textbf{84.07} \\ \hline
    \end{tabular}
}
\end{table}

\subsection{Results on DomainNet} \label{ssec:exp-domainnet}

Overall performance on DomainNet is summarized in Table \ref{tab:exp_total}, with detailed results reported in Table \ref{tab:domainnet_part1} and Table \ref{tab:domainnet_part2}.


\begin{table*}[tbh!]
	\caption{\textbf{Performance on DomainNet}, with clipart (c) and painting (p) as the source domain, respectively. Among the six UDA / UDE tasks, KDDE(DDC) tops the performance five times, followed by KDDE(CDAN). }
	\centering
	\label{tab:domainnet_part1}
	\scalebox{0.8}
	{
	\begin{tabular}{|l|c|c|c|c|c|c|c|c|c|}
	\hline
	\multirow{2}{*}{\textbf{Model}} & \multicolumn{3}{c|}{c$\rightarrow$p} & \multicolumn{3}{c|}{c$\rightarrow$r} & \multicolumn{3}{c|}{c$\rightarrow$s} \\ \cline{2-10} 
	 & c & p & c+p & c & r & c+r & c & s & c+s \\ \hline
	ResNet-50 & \textbf{77.16} & 32.07 & 54.62 & 77.16 & 48.22 & 62.69 & \textbf{77.16} & 38.50 & 57.83 \\ \hline
	DDC & 75.36 & 36.52 & 55.94 & 75.77 & 54.09 & 64.93 & 75.10 & 41.22 & 58.16 \\ \hline
	DANN & 71.35 & 33.51 & 52.43 & 73.92 & 52.98 & 63.45 & 73.45 & 40.41 & 56.93 \\ \hline
	CDAN & 72.45 & 34.33 & 53.39 & 73.34 & 53.23 & 63.29 & 72.25 & 39.08 & 55.67 \\ \hline
	KDDE(DDC) & 76.57 & \textbf{37.71} & \textbf{57.14} & 76.77 & 55.52 & 66.15 & 76.20 & \textbf{42.17} & \textbf{59.19} \\ \hline
	KDDE(CDAN) & 75.69 & 36.27 & 55.98 & \textbf{77.25} & \textbf{55.60} & \textbf{66.43} & 75.53 & 41.81 & 58.67 \\ \hline
	\multirow{2}{*}{\textbf{Model}} & \multicolumn{3}{c|}{p$\rightarrow$c} & \multicolumn{3}{c|}{p$\rightarrow$r} & \multicolumn{3}{c|}{p$\rightarrow$s} \\ \cline{2-10} 
	 & p & c & p+c & p & r & p+r & p & s & p+s \\ \hline
	ResNet-50 & \textbf{69.71} & 39.72 & 54.72 & 69.71 & 53.28 & 61.50 & \textbf{69.71} & 33.30 & 51.51 \\ \hline
	DDC & 65.40 & 44.86 & 55.13 & 68.98 & 58.48 & 63.73 & 65.01 & 37.93 & 51.47 \\ \hline
	DANN & 59.89 & 41.74 & 50.82 & 66.78 & 55.24 & 61.01 & 61.70 & 36.83 & 49.27 \\ \hline
	CDAN & 63.54 & 43.09 & 53.32 & 65.58 & 55.30 & 60.44 & 61.83 & 37.64 & 49.74 \\ \hline
	KDDE(DDC) & 67.45 & \textbf{46.73} & \textbf{57.09} & \textbf{70.39} & \textbf{59.91} & \textbf{65.15} & 66.20 & \textbf{39.60} & \textbf{52.90} \\ \hline
	KDDE(CDAN) & 66.34 & 45.07 & 55.71 & 69.68 & 57.64 & 63.66 & 65.19 & 39.53 & 52.36 \\ \hline
	\end{tabular}
	}
\end{table*}

\begin{table*}[tbh!]
	\caption{\textbf{Performance on DomainNet}, with real (r) and sketch (s) as the source domain, respectively. Among the six UDA tasks, both KDDE(DDC) and KDDE(CDAN) tops the performance three times. As for the six UDE tasks, KDDE(DDC) performs the best, followed by KDDE(CDAN).}
	\centering
	\label{tab:domainnet_part2}
	\scalebox{0.8}
	{
	\begin{tabular}{|l|c|c|c|c|c|c|c|c|c|}
	\hline
	\multirow{2}{*}{\textbf{Model}} & \multicolumn{3}{c|}{r$\rightarrow$c} & \multicolumn{3}{c|}{r$\rightarrow$p} & \multicolumn{3}{c|}{r$\rightarrow$s} \\ \cline{2-10} 
	 & r & c & r+c & r & p & r+p & r & s & r+s \\ \hline
	ResNet-50 & \textbf{82.96} & 49.60 & 66.28 & 82.96 & 45.71 & 64.34 & \textbf{82.96} & 34.50 & 58.73 \\ \hline
	DDC & 81.16 & 50.08 & 65.62 & 82.14 & 46.50 & 64.32 & 80.23 & 36.34 & 58.29 \\ \hline
	DANN & 77.25 & 49.32 & 63.29 & 78.34 & 43.25 & 60.80 & 76.85 & 37.84 & 57.35 \\ \hline
	CDAN & 79.10 & 50.99 & 65.05 & 80.63 & 46.30 & 63.47 & 78.03 & 40.02 & 59.03 \\ \hline
	KDDE(DDC) & 82.19 & 52.68 & 67.44 & \textbf{83.28} & 48.77 & \textbf{66.03} & 81.25 & 38.71 & 59.98 \\ \hline
	KDDE(CDAN) & 81.37 & \textbf{53.56} & \textbf{67.47} & 82.68 & \textbf{49.00} & 65.84 & 80.59 & \textbf{41.93} & \textbf{61.26} \\ \hline
	\multirow{2}{*}{\textbf{Model}} & \multicolumn{3}{c|}{s$\rightarrow$c} & \multicolumn{3}{c|}{s$\rightarrow$p} & \multicolumn{3}{c|}{s$\rightarrow$r} \\ \cline{2-10} 
	 & s & c & s+c & s & p & s+p & s & r & s+r \\ \hline
	ResNet-50 & \textbf{68.51} & 49.92 & 59.22 & \textbf{68.51} & 31.19 & 49.85 & 68.51 & 41.84 & 55.18 \\ \hline
	DDC & 66.57 & 54.26 & 60.42 & 66.48 & 41.15 & 53.82 & 67.04 & 52.97 & 60.01 \\ \hline
	DANN & 64.36 & 53.13 & 58.75 & 64.61 & 39.88 & 52.25 & 63.97 & 50.27 & 57.12 \\ \hline
	CDAN & 63.48 & 52.04 & 57.76 & 63.36 & 39.89 & 51.63 & 63.11 & 50.55 & 56.83 \\ \hline
	KDDE(DDC) & 68.33 & \textbf{56.46} & \textbf{62.40} & 68.15 & \textbf{43.47} & \textbf{55.81} & \textbf{68.54} & \textbf{54.75} & \textbf{61.65} \\ \hline
	KDDE(CDAN) & 66.57 & 55.34 & 60.96 & 66.70 & 42.51 & 54.61 & 68.19 & 53.51 & 60.85 \\ \hline
	\end{tabular}
	}
\end{table*}

\subsubsection{Performance on the target domain}

Similar to the results on Office-Home, we again observe that the domain adaptation models are effective for improving the performance of ResNet-50 on the target domain. In particular, as Table \ref{tab:exp_total} shows, DDC, CDAN and DANN obtain a relative improvement of 4.71\%, 3.72\% and 3.04\%, respectively. However, different from Office-Home, the classical DDC model now outperforms CDAN and DANN on DomainNet. The result suggests that although domain adaptation by adversarial learning is theoretically more appealing than the metric-based counterpart, much room exists for improving the adversarial methods. The proposed KDDE is again found to be effective, surpassing the best DDC with accuracy increased from 46.20 to 48.04, which accounts for a relative improvement of 3.98\%.


\subsubsection{Performance on the source domain}

The source-only ResNet-50 model performs expectedly best on the source domain. Compared to DDC, DANN and CDAN have relatively larger loss in performance. The result suggests that the adversarial methods perform domain adaptation more aggressively. With KDDE, the relative loss of DDC is reduced from 2.15\% to 0.81\%, and that of CDAN is reduced from 4.86\% to 1.61\%. This result again shows that KDDE better preserves the classification ability for the source domain. Recall that KDDE targets at the expanded domain, with knowledge from the source and target domain models treated equally. It is therefore less effective than ResNet50 exclusively trained on the source domain.



\subsubsection{Performance on the expanded domain}

The best baseline is DDC, obtaining a relative improvement of 1.28\% against ResNet-50. With KDDE, this number increases to 2.87\%. Moreover, KDDE consistently outperforms the baselines for all the 12 UDE tasks, see Table \ref{tab:domainnet_part1} and Table \ref{tab:domainnet_part2}. These results clearly justify the effectiveness of the proposed model for UDE. 






\begin{figure*}[tbh!]
	\setlength{\abovecaptionskip}{5pt}
	\subfigure[\textbf{clipart$\rightarrow$painting}]{
		\includegraphics[width=0.95\columnwidth]{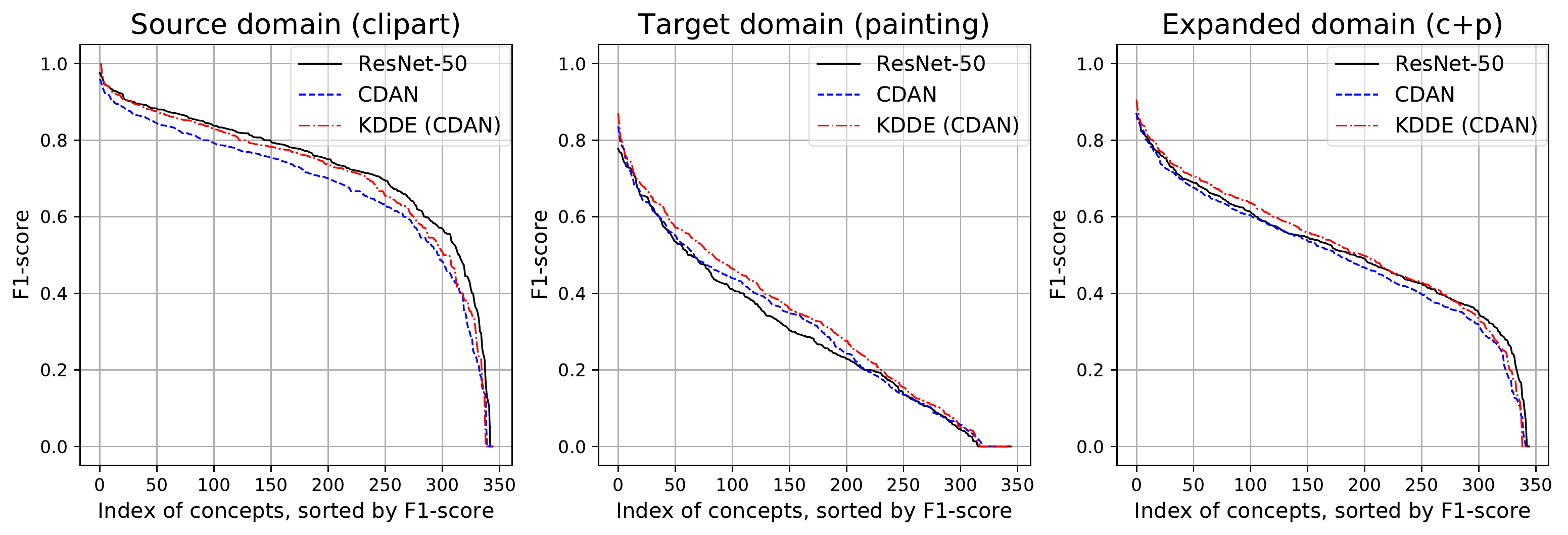}
		\label{fig:fscore-c2p}
	}
	\quad
	\subfigure[\textbf{clipart$\rightarrow$real}]{
		\includegraphics[width=0.95\columnwidth]{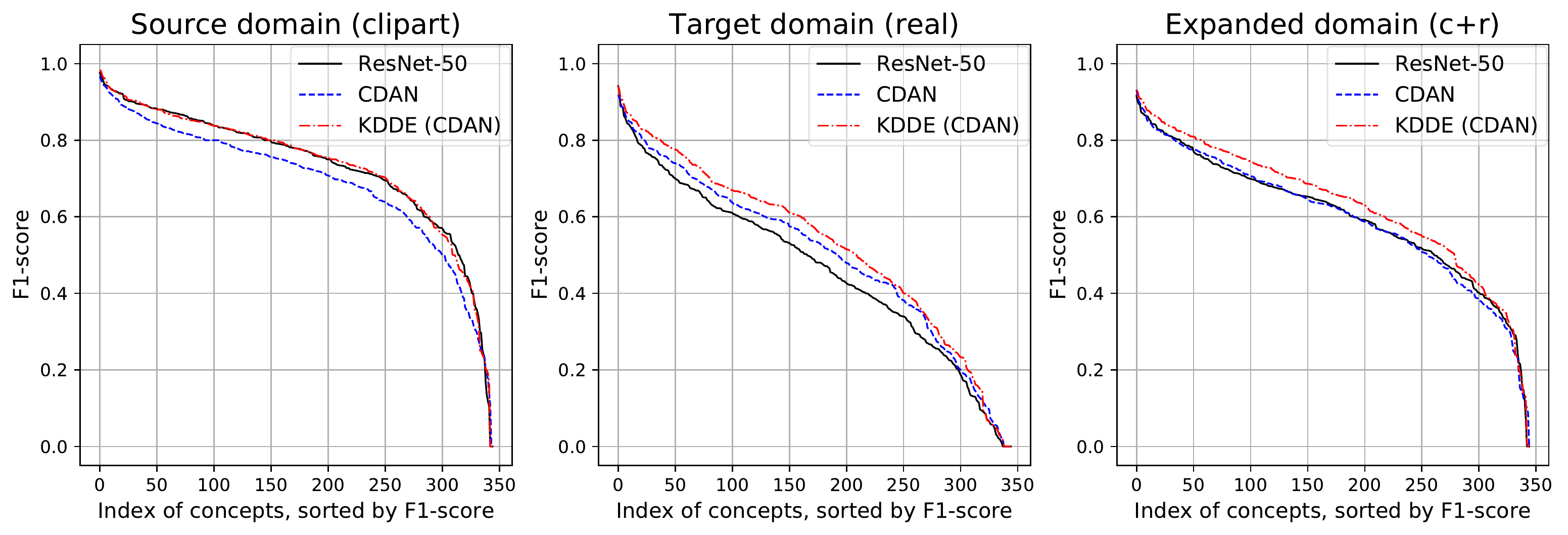}
		\label{fig:fscore-c2r}
	}
	\quad
	\subfigure[\textbf{clipart$\rightarrow$sketch}]{
		\includegraphics[width=0.95\columnwidth]{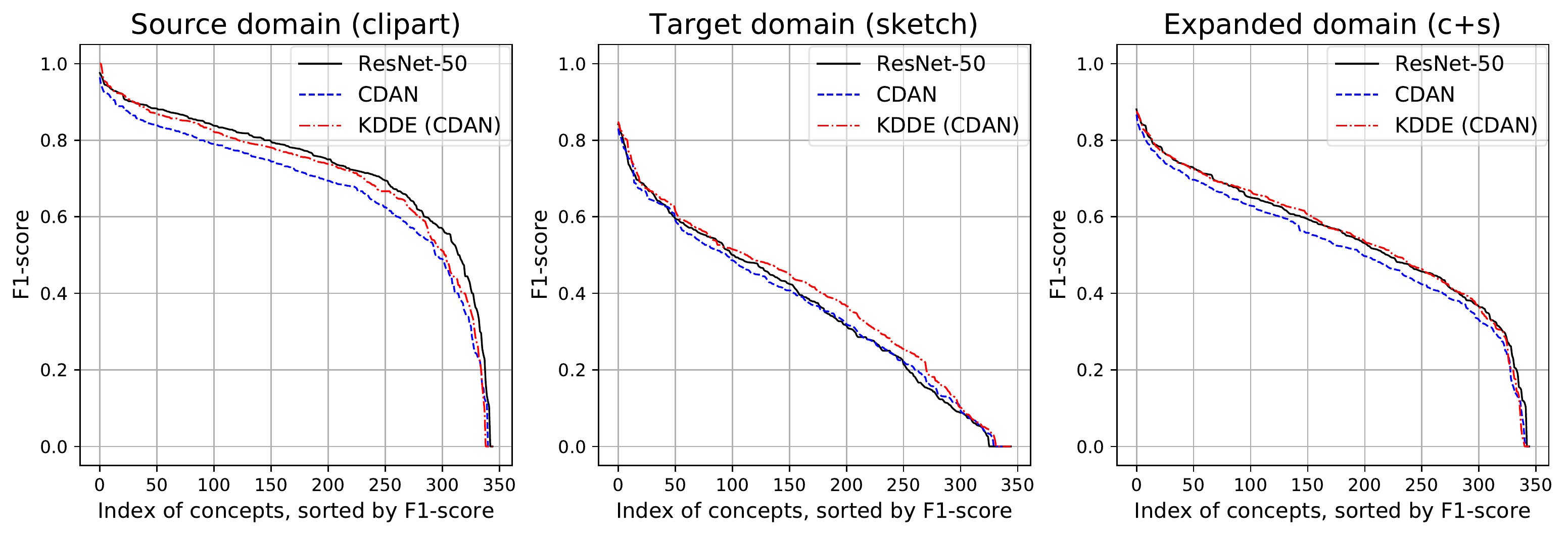}
	}
	\caption{\textbf{A concept-based comparison of ResNet-50, CDAN and KDDE (CDAN) on DomainNet}.  For the majority of the 345 concepts, KDDE(CDAN) is either better or comparable to the other two models on the target domain and on the expanded domain. }
	\label{fig:fscore}
\end{figure*}

Fig. \ref{fig:fscore} shows a concept-based comparison of ResNet-50, CDAN and KDDE(CDAN) on three UDE tasks, \ie clipart(c) $\rightarrow$ painting(p), clipart(c) $\rightarrow$ real(r) and clipart(c) $\rightarrow$ sketch(s). For the ease of comparison, for each model we have sorted all the 345 concepts in descending order according to their F1-scores. This allows us to measure the performance gap between the $j$-th best-performed concept of the models. As shown in the first column of Fig. \ref{fig:fscore}, when tested on the source domain, the $150$-th best F1-score of ResNet-50 and KDDE (CDAN) is approximately $0.8$, while the corresponding position of CDAN is noticeably lower. The curve of KDDE(CDAN) is between ResNet-50 and CDAN, indicating that the performance of CDAN for the source domain has been recovered to some extent by KDDE.
As for the target domain, see the middle column of Fig. \ref{fig:fscore}, KDDE(CDAN)is higher than CDAN, followed by ResNet-50, proving that KDDE(CDAN) is also beneficial for the UDA task. As shown in the last column, KDDE(CDAN) scores the best for the majority of the concepts on the expanded domain. 



\subsubsection{Qualitative analysis}

Fig. \ref{fig:tsne} presents the t-SNE \cite{TSNE} embedding of deep features learned by ResNet-50, CDAN and KDDE (CDAN) in the setting of clipart(c)$\rightarrow$painting(p). For better visualization, we only  show test examples of 10 concepts selected at random from the top 30 best-performed concepts of ResNet-50. Across the source, target and expanded domains, intra-class data points tend to stay closer while inter-class data points are more distant in the feature space of KDDE(CDAN).



In order to better understand the behavior of the three models, we further employ Grad-CAM~\cite{gradcam} to visualize how the decisions are made. Fig. \ref{fig:heatmaps} shows Grad-CAM based heatmaps, where the first three rows and the last three rows are test images selected from the source (clipart) and the target (real) domains, respectively. Note that ResNet-50 clearly differs from CDAN in terms of their salient areas. By contrast, KDDE(CDAN) imitates ResNet-50 on the source domain (first three rows), yet resembles CDAN on the target domain (last three rows). 
These heatmaps demonstrate how KDDE adaptively learns from the two domain-specific models.






\subsection{Experiments on OCT-11k}

Results on OCT-11k measured in terms of confusion matrices, accuracy and AUC are summarized in Table \ref{tab:oct}. Similar to our previous experiments on Office-Home and DomainNet, the source-only ResNet-50 performs well on the source domain (\emph{Zeiss}) but fails to generalize to the target domain (\emph{Topcon}). Its confusion matrix shows that a large number of 334 true negatives are incorrectly classified as positive. This is in line with its ROC curve in Fig. \ref{fig:oct-showcase}c that the operating point with default cutoff of 0.5 produces a relatively larger False Positive Rate (FPR). Interestingly, while the ROC curve of CDAN is quite close to that of ResNet-50, its operating point given the same cutoff obtains a smaller FPR. The results suggest that UDA effectively improves the model's insensitivity w.r.t. the value of the cutoff on the target domain. However, this advantage is obtained at the cost of noticeable performance drop on the source domain (0.8233 versus 0.8533 in accuracy and 0.8926 versus 0.9326 in AUC). By contrast, KDDE(CDAN) obtains the best overall performance, with no need of tuning the cutoff, justifying KDDE as a more principled approach to improving medical image classification in a multi-device scenario.

\begin{table}[thb!]
  \caption{\textbf{Performance of three ResNet-50 models on OCT-11k}, where ResNet-50 indicates the source-only model, followed by its counterparts trained by CDAN and KDDE(CDAN), respectively. Confusion matrices are shown in colored cells. KDDE(CDAN) is the best in terms of the overall performance.}
  \label{tab:oct}
  \resizebox{\textwidth}{!}{
  \begin{tabular}{|l|l|l|r|r|c|c|r|r|c|c|r|r|c|c|}
  \hline
   &
    \multicolumn{2}{c|}{} &
    \multicolumn{4}{c|}{\textbf{Source domain} (\emph{Zeiss})} &
    \multicolumn{4}{c|}{\textbf{Target domain} (\emph{Topcon})} &
    \multicolumn{4}{c|}{\textbf{Expanded domain} (\emph{Zeiss+Topcon})} \\ \cline{4-15} 
   &
    \multicolumn{2}{c|}{} &
    \multicolumn{2}{c|}{$y$} &
     &
     &
    \multicolumn{2}{c|}{$y$} &
     &
     &
    \multicolumn{2}{c|}{$y$} &
     &
     \\ \cline{4-5} \cline{8-9} \cline{12-13}
  \multirow{-3}{*}{\textbf{Model}} &
    \multicolumn{2}{c|}{\multirow{-3}{*}{}} &
    \multicolumn{1}{l|}{\textit{positive}} &
    \multicolumn{1}{l|}{\textit{negative}} &
    \multirow{-2}{*}{Accuracy} &
    \multirow{-2}{*}{AUC} &
    \multicolumn{1}{l|}{\textit{positive}} &
    \multicolumn{1}{l|}{\textit{negative}} &
    \multirow{-2}{*}{Accuracy} &
    \multirow{-2}{*}{AUC} &
    \multicolumn{1}{l|}{\textit{positive}} &
    \multicolumn{1}{l|}{\textit{negative}} &
    \multirow{-2}{*}{Accuracy} &
    \multirow{-2}{*}{AUC} \\ \hline
   &
     &
    \textit{positive} &
    \cellcolor[HTML]{0E59A2}{\color[HTML]{FFFFFF} 491} &
    \cellcolor[HTML]{DDEAF6}42 &
     &
     &
    \cellcolor[HTML]{083572}{\color[HTML]{FFFFFF} 462} &
    \cellcolor[HTML]{1B6AAF}{\color[HTML]{FFFFFF} 334} &
     &
     &
    \cellcolor[HTML]{08488F}{\color[HTML]{FFFFFF} 953} &
    \cellcolor[HTML]{6AADD5}376 &
     &
     \\ \cline{3-5} \cline{8-9} \cline{12-13}
  \multirow{-2}{*}{ResNet-50} &
    \multirow{-2}{*}{$\hat{y}$} &
    \textit{negative} &
    \cellcolor[HTML]{D8E7F5}90 &
    \cellcolor[HTML]{09539D}{\color[HTML]{FFFFFF} 277} &
    \multirow{-2}{*}{\textbf{0.8533}} &
    \multirow{-2}{*}{\textbf{0.9326}} &
    \cellcolor[HTML]{F3F8FD}9 &
    \cellcolor[HTML]{CBDEF0}95 &
    \multirow{-2}{*}{0.6189} &
    \multirow{-2}{*}{0.8593} &
    \cellcolor[HTML]{E4EFF9}99 &
    \cellcolor[HTML]{6DAFD6}372 &
    \multirow{-2}{*}{0.7361} &
    \multirow{-2}{*}{0.8534} \\ \hline
   &
     &
    \textit{positive} &
    \cellcolor[HTML]{1562A9}{\color[HTML]{FFFFFF} 471} &
    \cellcolor[HTML]{D8E7F5}49 &
     &
     &
    \cellcolor[HTML]{083A7A}{\color[HTML]{FFFFFF} 453} &
    \cellcolor[HTML]{75B3D8}204 &
     &
     &
    \cellcolor[HTML]{08519C}{\color[HTML]{FFFFFF} 924} &
    \cellcolor[HTML]{AACFE5}253 &
     &
     \\ \cline{3-5} \cline{8-9} \cline{12-13}
  \multirow{-2}{*}{CDAN} &
    \multirow{-2}{*}{$\hat{y}$} &
    \textit{negative} &
    \cellcolor[HTML]{D1E2F2}110 &
    \cellcolor[HTML]{0E59A2}{\color[HTML]{FFFFFF} 270} &
    \multirow{-2}{*}{0.8233} &
    \multirow{-2}{*}{0.8926} &
    \cellcolor[HTML]{EFF6FC}18 &
    \cellcolor[HTML]{63A9D3}225 &
    \multirow{-2}{*}{0.7533} &
    \multirow{-2}{*}{0.8526} &
    \cellcolor[HTML]{DEEBF7}128 &
    \cellcolor[HTML]{3989C1}495 &
    \multirow{-2}{*}{0.7883} &
    \multirow{-2}{*}{0.8439} \\ \hline
   &
     &
    \textit{positive} &
    \cellcolor[HTML]{0E59A2}{\color[HTML]{FFFFFF} 492} &
    \cellcolor[HTML]{DBE9F6}45 &
     &
     &
    \cellcolor[HTML]{084388}{\color[HTML]{FFFFFF} 437} &
    \cellcolor[HTML]{A8CEE4}148 &
     &
     &
    \cellcolor[HTML]{084F99}{\color[HTML]{FFFFFF} 929} &
    \cellcolor[HTML]{C4DAEE}193 &
     &
     \\ \cline{3-5} \cline{8-9} \cline{12-13}
  \multirow{-2}{*}{KDDE(CDAN)} &
    \multirow{-2}{*}{$\hat{y}$} &
    \textit{negative} &
    \cellcolor[HTML]{D8E7F5}89 &
    \cellcolor[HTML]{0B559F}{\color[HTML]{FFFFFF} 274} &
    \multirow{-2}{*}{0.8511} &
    \multirow{-2}{*}{0.9200} &
    \cellcolor[HTML]{E8F1FA}34 &
    \cellcolor[HTML]{3A8AC1}281 &
    \multirow{-2}{*}{\textbf{0.7978}} &
    \multirow{-2}{*}{\textbf{0.8985}} &
    \cellcolor[HTML]{E0ECF7}123 &
    \cellcolor[HTML]{2373B6}{\color[HTML]{FFFFFF} 555} &
    \multirow{-2}{*}{\textbf{0.8244}} &
    \multirow{-2}{*}{\textbf{0.8914}} \\ \hline
  \end{tabular}
  }
\end{table}



\subsection{Ablation Study}

\subsubsection{Alternative knowledge distillation loss?}\label{sssec:kd-loss} 

We compare the KL divergence loss with two alternatives, \ie the $l_2$ loss and the cross-entropy loss. As shown in Table \ref{tab:kd-loss}, the KL divergence loss performs the best. 


\begin{table}[tbh!]
	\caption{\textbf{Performance of KDDE (CDAN) with different losses on Office-Home}.}
	\label{tab:kd-loss}
	\scalebox{0.8}{
	\begin{tabular}{|l|c|c|c|}
	\hline
	\textbf{Loss} & \textbf{Source domain} & \textbf{Target domain} & \textbf{Expanded domain} \\ 
	\hline
	$l_2$ & 79.60 & 61.77 & 70.69 \\ 
	\hline
	\textit{cross-entropy} & 80.78 & 63.19 & 71.99 \\ 
	\hline
	\textit{KL divergence} & 81.44 & \textbf{63.90} & \textbf{72.67} \\ 
	\hline
	\end{tabular}
	}
\end{table}

\subsubsection{The effect of knowledge distillation at the intermediate features space} \label{sssec:feat-space}

\begin{figure*}[tbh!]
	\setlength{\abovecaptionskip}{5pt}
	\centering
	\includegraphics[width=\textwidth]{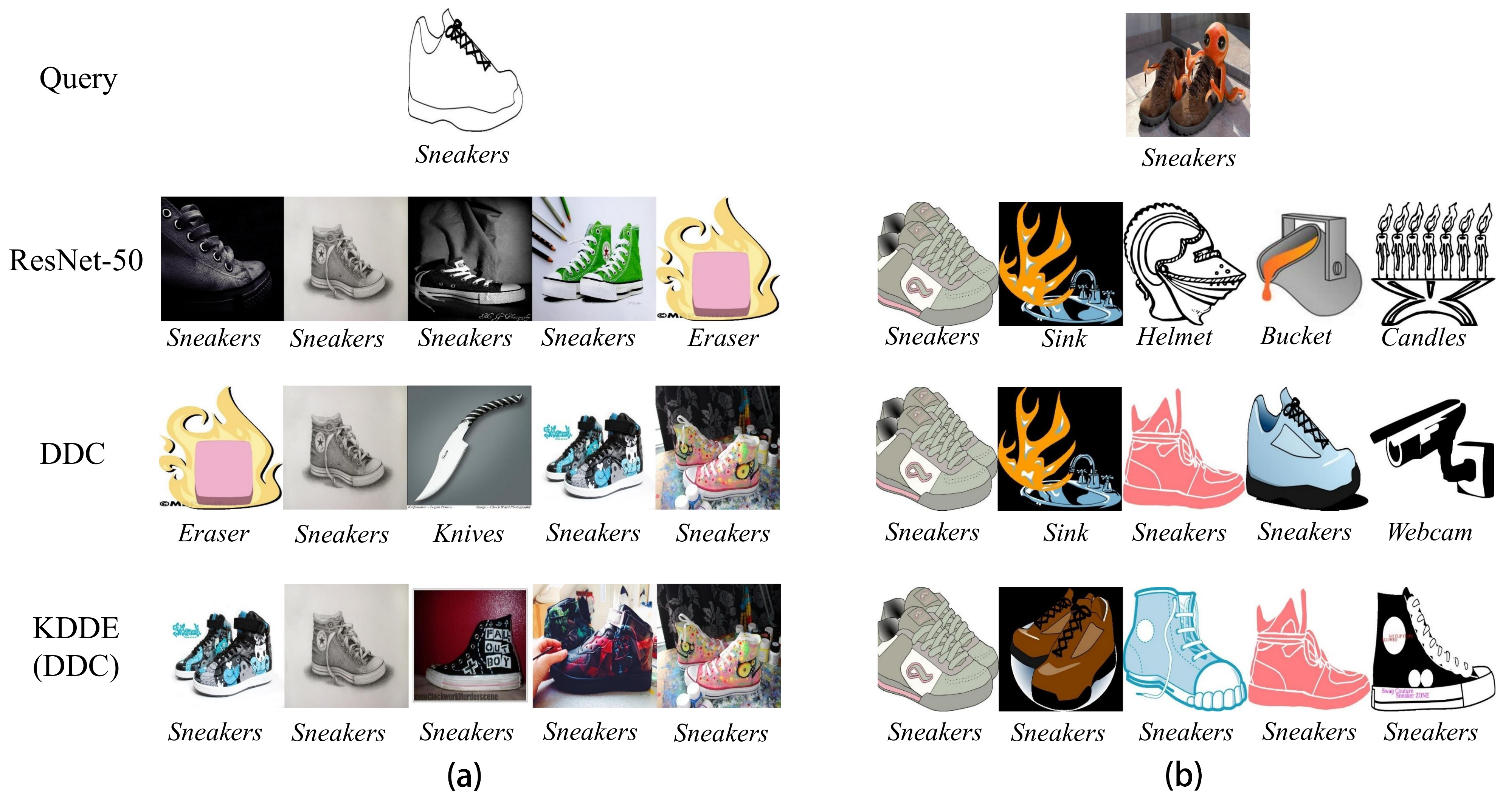}
	\caption{\textbf{Visualization of cross-domain image retrieval}. Each row corresponds to the top-5 retrieved images for queries from (a) the \textit{Clipart} domain and (b) the \textit{Art} domain, respectively. Note that image retrieval is fully content-based. Labels below each image are for illustration only.}
	\label{fig:cross-domain}
\end{figure*}

Features that are discriminative of both domains shall allow an instance to be surrounded by instances of the same class. In order to verify if features obtained by KDDE are more discriminative, we consider cross-domain image retrieval, where each test image from one domain is used as a query example to retrieve images from the other domain. In particular, we conduct cross-domain image retrieval on two UDE tasks, \ie C$\rightarrow$A and P$\rightarrow$R, on Office-Home. We compare ResNet-50, DDC and KDDE(DDC). Per model, the dissimilarity between two images $x$ and $x'$ is defined as the $l_2$ distance between their 2,048-d features $F(x)$ and $F(x')$. 
As Table \ref{tab:cross-domain} shows, using features of KDDE obtains higher precisions, indicating that cross-domain instances of the same class stay more closer in the intermediate feature space. Some qualitative results are presented in Fig. \ref{fig:cross-domain}, where the top-5 returned items w.r.t. KDDE consistently exhibit domain-invariant visual patterns of sneakers. Both quantitative and qualitative results allow us to conclude that knowledge distillation results in more discriminative and domain-invariant feature representations.


\begin{table}[tbh!]
	\caption{\textbf{Performance of cross-domain image retrieval}, using the 2,048-d features produced by the feature extractor $F$ of ResNet-50, DDC and KDDE (DDC), respectively. ``\textit{Query} \textit{S} on \textit{T}'' means using each image in a source domain as a query example and performing retrieval in the target domain, while ``\textit{Query} \textit{T} on \textit{S}'' indicates retrieval in the opposite direction. Performance metric: Precision at $N$ (P@N), $N=5,10$. }
	\label{tab:cross-domain}
	\centering
	\scalebox{0.8}{
		\begin{tabular}{|l|c|c|c|c|c|c|c|c|}
		\hline
		\multicolumn{1}{|c|}{\multirow{3}{*}{\textbf{Model}}} & \multicolumn{4}{c|}{C $\rightarrow$ A}                                & \multicolumn{4}{c|}{P $\rightarrow$ R}                                \\ \cline{2-9} 
		\multicolumn{1}{|c|}{}                       & \multicolumn{2}{c|}{\textit{Query} \textit{S} on \textit{T}} & \multicolumn{2}{c|}{\textit{Query} \textit{T} on \textit{S}} & \multicolumn{2}{c|}{\textit{Query} \textit{S} on \textit{T}} & \multicolumn{2}{c|}{\textit{Query} \textit{T} on \textit{S}} \\ \cline{2-9} 
		\multicolumn{1}{|c|}{}                       & P@5            & P@10           & P@5            & P@10           & P@5            & P@10           & P@5            & P@10           \\ \hline
		ResNet-50                                    & 47.87           & 41.09           & 40.24           & 38.07           & 77.67           & 74.13           & 65.19           & 63.26           \\ \hline
		DDC                                          & 43.43           & 36.20           & 43.86           & 40.75           & 72.83           & 68.23           & 64.95           & 61.86           \\ \hline
		KDDE(DDC)                                    & \textbf{49.86}  & \textbf{42.98}  & \textbf{47.29}  & \textbf{45.08}  & \textbf{78.23}  & \textbf{74.61}  & \textbf{68.10}  & \textbf{66.38}  \\ \hline
		\end{tabular}
	}
\end{table}
	
\subsubsection{KDDE for multi-domain shifts} \label{sssec:eval-domain-shifts}

Existing works on domain adaptation typically consider single-domain shift, where one wants to adapt a model for a single target domain, \eg A$\rightarrow C$. We investigate a more challenging scenario of multi-domain shifts, which is to generalize the model simultaneously to multiple target domains, \eg A$\rightarrow$ \{C, P, R\}.
To that end, we improve CDAN by modifying its binary domain discriminator to support 4-way classification. We term the variant CDAN+. The performance of CDAN+ and KDDE (CDAN+) is reported in Table \ref{tab:multi-domain}. Patterns similar to the previous single-domain shift experiments are observed. That is, CDAN+ outperforms the source-only ResNet-50 on the target domains but is less effective on the original source domain, while KDDE is better than CDAN+ on both source and multiple target domains. Consequently, KDDE obtains the overall best performance on the expanded domain. 

\begin{table}[thb!]
    \caption{\textbf{Performance of multi-domain shifts on Office-Home}. A$\rightarrow$\{C,P,R\} means using A as a source domain, which will be expanded to cover C, P and R. CDAN+ is our improved version of CDAN~\cite{nips18-cdan} that performs 4-way classification in its domain discriminator. The source-only ResNet-50 is used as a reference. Performance increases (decreases) over this reference are shown in red (green).}
    \label{tab:multi-domain}
    \scalebox{0.8}{
    \begin{tabular}{|l|l|l|l|l|l|}
    \hline
    \multicolumn{1}{|c|}{\multirow{2}{*}{\textbf{Model}}} & \multicolumn{5}{c|}{A$\rightarrow$\{C,P,R\}} \\ \cline{2-6} 
    \multicolumn{1}{|c|}{}      & A                                            & C                                              & P                                     & R                                            & A+C+P+R                              \\ \hline
    ResNet-50                   & \textbf{75.20}                               & 45.23                                          & 58.45                                 & 69.35                                        & 62.06                                \\ \hline
    CDAN+                       & 70.64$\downarrow$\textcolor{darkgreen}{-4.56} & \textbf{50.23}$\uparrow$\textcolor{red}{+5.00}           & 61.35$\uparrow$\textcolor{red}{+2.90}  & 67.17$\downarrow$\textcolor{darkgreen}{-2.18} & 62.35$\uparrow$\textcolor{red}{+0.29} \\ \hline
    KDDE(CDAN+)                 & 72.59$\downarrow$\textcolor{darkgreen}{-2.61} & 50.14$\uparrow$\textcolor{red}{+4.91}           & \textbf{63.49}$\uparrow$\textcolor{red}{+5.04}  & \text{69.49}$\uparrow$\textcolor{red}{+0.14}         & \textbf{63.93}$\uparrow$\textcolor{red}{+1.87} \\ \hline

    \multicolumn{1}{|c|}{\multirow{2}{*}{\textbf{Model}}} & \multicolumn{5}{c|}{C$\rightarrow$\{A,P,R\}} \\ \cline{2-6} 
    \multicolumn{1}{|c|}{}      & C                                            & A                                              & P                                     & R                                            & C+A+P+R                              \\ \hline
    ResNet-50                   & 78.91                                        & 47.06                                          & 57.55                                 & 59.65                                        & 60.79                                \\ \hline
    CDAN+                       & 77.09$\downarrow$\textcolor{darkgreen}{-1.82} & 54.89$\uparrow$\textcolor{red}{+7.83}           & 66.53$\uparrow$\textcolor{red}{+8.98}  & 64.62$\uparrow$\textcolor{red}{+4.97}         & 65.78$\uparrow$\textcolor{red}{+4.99} \\ \hline
    KDDE(CDAN+)                 & \textbf{79.05}$\uparrow$\textcolor{red}{+0.14}         & \textbf{57.67}$\uparrow$\textcolor{red}{+10.61} & \textbf{68.19}$\uparrow$\textcolor{red}{+10.64} & \textbf{66.94}$\uparrow$\textcolor{red}{+7.29}         & \textbf{67.96}$\uparrow$\textcolor{red}{+7.17} \\ \hline

    \multicolumn{1}{|c|}{\multirow{2}{*}{\textbf{Model}}} & \multicolumn{5}{c|}{P$\rightarrow$\{A,C,R\}} \\ \cline{2-6} 
    \multicolumn{1}{|c|}{}      & P                                            & A                                              & C                                     & R                                            & P+A+C+R                              \\ \hline
    ResNet-50                   & \textbf{92.05}                               & 50.33                                          & 43.86                                 & 70.31                                        & 64.14                                \\ \hline
    CDAN+                       & 90.08$\downarrow$\textcolor{darkgreen}{-1.97} & 54.24$\uparrow$\textcolor{red}{+3.91}           & 47.91$\uparrow$\textcolor{red}{+4.05}  & 70.90$\uparrow$\textcolor{red}{+0.59}         & 65.78$\uparrow$\textcolor{red}{+1.64} \\ \hline
    KDDE(CDAN+)                 & 91.33$\downarrow$\textcolor{darkgreen}{-0.72} & \text{56.28}$\uparrow$\textcolor{red}{+5.95}           & \textbf{50.50}$\uparrow$\textcolor{red}{+6.64}  & \textbf{72.04}$\uparrow$\textcolor{red}{+1.73}         & \textbf{67.54}$\uparrow$\textcolor{red}{+3.40}  \\ \hline

    \multicolumn{1}{|c|}{\multirow{2}{*}{\textbf{Model}}} & \multicolumn{5}{c|}{R$\rightarrow$\{A,C,P\}} \\ \cline{2-6} 
    \multicolumn{1}{|c|}{}      & R                                            & A                                              & C                                     & P                                            & R+A+C+P                              \\ \hline
    ResNet-50                   & \textbf{84.11}                               & \textbf{63.62}                                 & 48.23                                 & 76.27                                        & 68.06                                \\ \hline
    CDAN+                       & 82.01$\downarrow$\textcolor{darkgreen}{-2.10} & 62.40$\downarrow$\textcolor{darkgreen}{-1.22}   & 57.14$\uparrow$\textcolor{red}{+8.91}  & 76.32$\uparrow$\textcolor{red}{+0.05}         & 69.47$\uparrow$\textcolor{red}{+1.41} \\ \hline
    KDDE(CDAN+)                 & 82.65$\downarrow$\textcolor{darkgreen}{-1.46} & 63.30$\downarrow$\textcolor{darkgreen}{-0.32}   & \textbf{57.27}$\uparrow$\textcolor{red}{+9.04}  & \textbf{77.97}$\uparrow$\textcolor{red}{+1.70}          & \textbf{70.30}$\uparrow$\textcolor{red}{+2.24} \\ \hline
    \end{tabular}
    }
\end{table}

\subsubsection{Tackling UDE by tuning the trade-off parameter $\lambda$?} \label{sssec:lambda-effect}

We report in Table \ref{tab:different-lambdas} performance of DDC with the trade-off parameter $\lambda$ chosen from \{0, 0.01, 0.1, 1, 10, 20, 100\}, where $\lambda=10$ is the choice we have used so far. The peak performance is reached given $\lambda=0.01$, yet remains lower than KDDE. Moreover, by highlighting the best-performed $\lambda$ per domain using light-blue cells, we see that its optimal value is domain-dependent. Compared to simply tunning the trade-off parameter, KDDE is a more principled approach.


It is worth mentioning that DDC ($\lambda$=0) is not equivalent to the source-only ResNet-50. Because mini-batches sampled from the target domain have been used together with source mini-batches to estimate mean and variance of the Batch Normalization (BN) layers. This improves domain-variance of the intermediate features to some extent, and consequently leads to better performance (71.59 \textit{versus} 70.11 in Table \ref{tab:different-lambdas}).

\begin{table}[tbh!]
    \caption{\textbf{Performance of DDC w.r.t. the trade-off parameter $\lambda$}, on three tasks (C$\rightarrow$A, P$\rightarrow$R and R$\rightarrow$C) of Office-Home. Varied positions of light-blue cells, which indicate the best-performed $\lambda$ on specific domains, show the difficulty in selecting a proper $\lambda$ for tackling the UDE tasks.}
	\label{tab:different-lambdas}
	\centering
	\scalebox{0.75}{
		\begin{tabular}{|c|l|c|c|c|c|c|c|c|c|c|c|}
			\hline
			\multicolumn{2}{|c|}{\multirow{2}{*}{\textbf{Model}}} & \multicolumn{3}{c|}{C$\rightarrow$A}                                                                            & \multicolumn{3}{c|}{P$\rightarrow$R}                                                                   & \multicolumn{3}{c|}{R$\rightarrow$C}                                                          & \multirow{2}{*}{\begin{tabular}[c]{@{}c@{}}\textbf{Averaged accuracy}\\ \textbf{on expanded domains} \end{tabular}} \\ \cline{3-11}
			\multicolumn{2}{|c|}{}                       & C                                   & A                                   & C+A                                 & P                          & R                                   & P+R                                 & R                          & C                                   & R+C                        &                                                                                                \\ \hline
			\multicolumn{2}{|c|}{ResNet-50}              & 78.91                               & 47.06                               & 62.99                               & 92.05                      & 70.31                               & 81.18                               & \textbf{84.11}             & 48.23                               & 66.17                      & 70.11                                                                                          \\ \hline
			\multicolumn{2}{|l|}{KDDE(DDC, $\lambda$=10)} & \multicolumn{1}{l|}{\textbf{80.05}} & \multicolumn{1}{l|}{\textbf{54.57}} & \multicolumn{1}{l|}{\textbf{67.31}} & \multicolumn{1}{l|}{92.67} & \multicolumn{1}{l|}{\textbf{73.36}} & \multicolumn{1}{l|}{\textbf{83.02}} & \multicolumn{1}{l|}{83.24} & \multicolumn{1}{l|}{\textbf{53.86}} & \multicolumn{1}{l|}{68.55} & \textbf{72.96}                                                                                 \\ \hline
			\multirow{7}{*}{DDC}     & $\lambda$=0       & 78.68                               & 51.55                               & 65.12                               & 92.05                      & 71.54                               & 81.80                               & 83.01                      & 52.73                               & 67.87                      & 71.59                                                                                          \\ \cline{2-12} 
									 & $\lambda$=0.01    & 79.41                               & 52.28                               & 65.85                               & 92.40                      & 72.54                               & 82.47                               & \cellcolor[HTML]{8CC5F8}83.56   & \cellcolor[HTML]{8CC5F8}53.82  & \cellcolor[HTML]{8CC5F8}{\textbf{68.69}}  & 72.34                                                    \\ \cline{2-12} 
									 & $\lambda$=0.1     & 79.86                               & \cellcolor[HTML]{8CC5F8}52.37       & \cellcolor[HTML]{8CC5F8}66.12       & 92.58                      & 72.13                               & 82.36                               & 82.97                      & 53.77                               & 68.37                      & 72.28                                                                                          \\ \cline{2-12} 
									 & $\lambda$=1       & 79.23                               & 51.47                               & 65.35                               & \cellcolor[HTML]{8CC5F8}{\textbf{92.72} }  & \cellcolor[HTML]{8CC5F8}73.04 & \cellcolor[HTML]{8CC5F8}82.88 & 83.42                  & 52.82                               & 68.12                      & 72.12                                                                                          \\ \cline{2-12} 
									 & $\lambda$=10      & \cellcolor[HTML]{8CC5F8}80.23       & 50.90                               & 65.57                               & 92.49                      & 71.95                               & 82.22                               & 82.97                      & 52.23                               & 67.60                      & 71.80                                                                                          \\ \cline{2-12} 
									 & $\lambda$=20      & 79.23                               & 51.47                               & 65.35                               & 92.09                      & 71.27                               & 81.68                               & 83.20                      & 52.14                               & 67.67                      & 71.57                                                                                          \\ \cline{2-12} 
									 & $\lambda$=100     & 7.96                                & 2.53                                & 5.25                                & 13.50                      & 5.64                                & 9.57                                & 6.33                       & 3.41                                & 4.87                       & 6.59                                                                                           \\ \hline
			\end{tabular}
	}
\end{table}

\subsubsection{KDDE with MCD} \label{sssec:eval-mcd}

As we have noted in Section \ref{sec:method}, the UDA module of the proposed KDDE method can be implemented using any state-of-the-art UDA model. Here we instantiate the module using MCD~\cite{cvpr2018-mcd}. Again, ResNet-50 is used as their backbones. As shown in Table \ref{tab:MCD}, KDDE(MCD) consistently outperforms MCD.


\begin{table}[tbh!]
    \centering
    \caption{\textbf{Performance of MCD and KDDE (MCD)}. Per model we run the experiment three times, reporting the average value and standard deviation of the resultant accuracy scores.} 
    \label{tab:MCD}
    \scalebox{0.8}{
    \begin{tabular}{|c|c|c|c|c|c|c|}
    \hline
    \multirow{2}{*}{\textbf{Model}} & \multicolumn{3}{c|}{C$\rightarrow$A}           & \multicolumn{3}{c|}{P$\rightarrow$R}           \\ \cline{2-7} 
                           & C              & A              & C+A            & P              & R              & P+R            \\ \hline
    MCD                    & 77.53$\pm$0.66 & 51.90$\pm$0.96 & 64.72$\pm$0.53 & 91.45$\pm$0.32 & 70.67$\pm$0.71 & 81.06$\pm$0.51 \\ \hline
    KDDE(MCD)              & \textbf{79.20}$\pm$0.35 & \textbf{56.33}$\pm$0.65 & \textbf{67.77}$\pm$0.49 & \textbf{92.25}$\pm$0.16 & \textbf{73.91}$\pm$0.10 & \textbf{83.08}$\pm$0.04 \\ \hline
    \end{tabular}
    }
\end{table}

\subsubsection{Reproducibility test} \label{sssec:eval-reprod}

Due to the large-scale datasets used in our study, producing Table \ref{tab:exp_total} alone needs around 750 GPU hours, when running the related experiments in parallel on four GPU cards (two GTX 2080Ti plus two 1080Ti GPUs). Probably because of this reason, we rarely see reproducibility test in the literature of domain adaptation. Nonetheless, to reduce randomness, we run the experiments three times for all the tasks of Office-Home. As Table \ref{tab:repeat-experiments} shows, our major conclusion, \ie the proposed KDDE is more effective, is again confirmed. 


\begin{table}[tbh!]
    \setlength{\belowcaptionskip}{-5pt}
    \centering
    \caption{\textbf{Averaged performance of the individual models on Office-Home}. Per model we repeat the training and evaluation procedures three times, reporting the averaged accuracy and standard deviation.} 
    \label{tab:repeat-experiments}
    \scalebox{0.65}
    {
    \begin{tabular}{|c|c|c|c|c|c|c|c|c|c|}
    \hline
        \multirow{2}{*}{\textbf{Model}} & \multicolumn{3}{c|}{A$\rightarrow$C}                                        & \multicolumn{3}{c|}{A$\rightarrow$P}                                        & \multicolumn{3}{c|}{A$\rightarrow$R}                                        \\ \cline{2-10} 
    \multicolumn{1}{|l|}{}                                & A                       & C                       & A+C                     & A                       & P                       & A+P                     & A                       & R                       & A+R                     \\ \hline
    \multicolumn{1}{|l|}{ResNet-50}                       & \textbf{74.64$\pm$0.50} & 44.73$\pm$0.45          & 59.68$\pm$0.48          & \textbf{74.64$\pm$0.50} & 59.19$\pm$0.70          & 66.91$\pm$0.11          & 74.64$\pm$0.50          & 69.17$\pm$0.24          & 71.91$\pm$0.33          \\ \hline
    \multicolumn{1}{|l|}{DDC}                             & 73.03$\pm$0.58          & 48.58$\pm$0.48          & 60.80$\pm$0.27          & 73.60$\pm$0.17          & 62.99$\pm$0.33          & 68.29$\pm$0.17          & 74.31$\pm$0.35          & 70.52$\pm$0.31          & 72.42$\pm$0.04          \\ \hline
    \multicolumn{1}{|l|}{DANN}                            & 71.94$\pm$0.75          & \textbf{49.58$\pm$0.34} & 60.76$\pm$0.54          & 72.92$\pm$0.41          & 62.15$\pm$1.02          & 67.54$\pm$0.33          & 74.71$\pm$0.45          & 70.60$\pm$0.38          & 72.66$\pm$0.41          \\ \hline
    \multicolumn{1}{|l|}{DAAN}                            & 73.68$\pm$0.33          & 49.00$\pm$0.08          & 61.34$\pm$0.18          & 74.28$\pm$0.29          & 63.93$\pm$0.33          & 69.10$\pm$0.04          & 74.74$\pm$0.48          & 71.25$\pm$0.30          & 73.00$\pm$0.38          \\ \hline
    \multicolumn{1}{|l|}{CDAN}                            & 70.06$\pm$1.10          & 47.15$\pm$1.37          & 58.61$\pm$1.07          & 70.80$\pm$0.91          & 64.64$\pm$2.10          & 67.72$\pm$1.28          & 72.57$\pm$0.69          & 69.63$\pm$0.87          & 71.10$\pm$0.59          \\ \hline
    \multicolumn{1}{|l|}{KDDE(DDC)}                       & 73.38$\pm$0.26          & 49.50$\pm$0.62          & \textbf{61.44$\pm$0.42} & 74.33$\pm$0.66          & 64.57$\pm$0.79          & \textbf{69.45$\pm$0.45} & \textbf{75.58$\pm$0.58} & \textbf{71.84$\pm$0.56} & \textbf{73.71$\pm$0.57} \\ \hline
    \multicolumn{1}{|l|}{KDDE(CDAN)}                      & 68.84$\pm$1.08          & 47.98$\pm$0.84          & 58.41$\pm$0.94          & 71.53$\pm$0.80          & \textbf{66.20$\pm$1.93} & 68.87$\pm$1.32          & 73.68$\pm$0.29          & 71.19$\pm$1.14          & 72.43$\pm$0.71          \\ \hline
    \multirow{2}{*}{\textbf{Model}}                       & \multicolumn{3}{c|}{C$\rightarrow$A}                                        & \multicolumn{3}{c|}{C$\rightarrow$P}                                        & \multicolumn{3}{c|}{C$\rightarrow$R}                                        \\ \cline{2-10} 
                                                          & C                       & A                       & C+A                     & C                       & P                       & C+P                     & C                       & R                       & C+R                     \\ \hline
    \multicolumn{1}{|l|}{ResNet-50}                       & 78.98$\pm$0.09          & 48.20$\pm$1.28          & 63.59$\pm$0.65          & 78.98$\pm$0.09          & 58.31$\pm$0.69          & 68.65$\pm$0.37          & 78.98$\pm$0.09          & 59.85$\pm$0.35          & 69.42$\pm$0.17          \\ \hline
    \multicolumn{1}{|l|}{DDC}                             & 79.74$\pm$0.44          & 51.93$\pm$0.91          & 65.84$\pm$0.24          & 80.03$\pm$0.38          & 61.75$\pm$0.74          & 70.89$\pm$0.18          & 79.53$\pm$0.66          & 64.01$\pm$0.51          & 71.77$\pm$0.56          \\ \hline
    \multicolumn{1}{|l|}{DANN}                            & 78.06$\pm$0.23          & 53.10$\pm$0.91          & 65.58$\pm$0.52          & 78.86$\pm$0.69          & 60.41$\pm$1.24          & 69.64$\pm$0.82          & 78.89$\pm$0.03          & 62.93$\pm$0.33          & 70.91$\pm$0.15          \\ \hline
    \multicolumn{1}{|l|}{DAAN}                            & 79.08$\pm$0.48          & 52.80$\pm$0.74          & 65.94$\pm$0.43          & 79.51$\pm$0.45          & 62.21$\pm$0.34          & 70.87$\pm$0.19          & 79.74$\pm$0.07          & 64.51$\pm$0.41          & 72.13$\pm$0.19          \\ \hline
    \multicolumn{1}{|l|}{CDAN}                            & 78.09$\pm$0.30          & 53.59$\pm$1.28          & 65.84$\pm$0.59          & 78.27$\pm$0.64          & 65.34$\pm$0.69          & 71.81$\pm$0.33          & 79.12$\pm$0.14          & 64.66$\pm$0.64          & 71.89$\pm$0.33          \\ \hline
    \multicolumn{1}{|l|}{KDDE(DDC)}                       & \textbf{80.09$\pm$0.48} & 55.57$\pm$0.89          & \textbf{67.83$\pm$0.47} & \textbf{80.43$\pm$0.33} & 64.06$\pm$0.79          & 72.25$\pm$0.53          & 80.47$\pm$0.42          & \textbf{66.64$\pm$1.34} & \textbf{73.55$\pm$0.85} \\ \hline
    \multicolumn{1}{|l|}{KDDE(CDAN)}                      & 78.92$\pm$0.77          & \textbf{56.09$\pm$1.72} & 67.51$\pm$0.85          & 80.03$\pm$0.23          & \textbf{67.55$\pm$1.00} & \textbf{73.79$\pm$0.39} & \textbf{80.48$\pm$0.29} & 66.12$\pm$0.66          & 73.30$\pm$0.41          \\ \hline
    \multirow{2}{*}{\textbf{Model}}                       & \multicolumn{3}{c|}{P$\rightarrow$A}                                        & \multicolumn{3}{c|}{P$\rightarrow$C}                                        & \multicolumn{3}{c|}{P$\rightarrow$R}                                        \\ \cline{2-10} 
                                                          & P                       & A                       & P+A                     & P                       & C                       & P+C                     & P                       & R                       & P+R                     \\ \hline
    \multicolumn{1}{|l|}{ResNet-50}                                             & 92.05$\pm$0.23          & 52.20$\pm$1.63          & 72.13$\pm$0.81          & \textbf{92.05$\pm$0.23} & 42.94$\pm$0.89          & 67.49$\pm$0.50          & 92.05$\pm$0.23          & 70.11$\pm$0.34          & 81.08$\pm$0.09          \\ \hline
    \multicolumn{1}{|l|}{DDC}                                                   & 92.20$\pm$0.14          & 52.99$\pm$0.62          & 72.59$\pm$0.29          & 91.69$\pm$0.12          & 45.39$\pm$0.42          & 68.54$\pm$0.25          & 92.30$\pm$0.20          & 72.42$\pm$0.44          & 82.36$\pm$0.12          \\ \hline
    \multicolumn{1}{|l|}{DANN}                                                  & 90.32$\pm$0.36          & 51.55$\pm$1.56          & 70.93$\pm$0.94          & 90.28$\pm$0.49          & 47.52$\pm$0.43          & 68.90$\pm$0.36          & 91.76$\pm$0.12          & 71.56$\pm$0.90          & 81.66$\pm$0.39          \\ \hline
    \multicolumn{1}{|l|}{DAAN}                                                  & 92.06$\pm$0.50          & \textbf{54.21$\pm$0.87} & \textbf{73.14$\pm$0.37} & 91.63$\pm$0.24          & 45.24$\pm$1.05          & 68.44$\pm$0.61          & 92.38$\pm$0.21          & 72.37$\pm$0.22          & 82.38$\pm$0.18          \\ \hline
    \multicolumn{1}{|l|}{CDAN}                                                  & 90.44$\pm$0.59          & 52.37$\pm$1.14          & 71.40$\pm$0.86          & 89.11$\pm$0.49          & 48.33$\pm$1.27          & 68.72$\pm$0.40          & 90.97$\pm$0.20          & 73.98$\pm$0.41          & 82.48$\pm$0.18          \\ \hline
    \multicolumn{1}{|l|}{KDDE(DDC)}                                             & \textbf{92.07$\pm$0.28} & 54.16$\pm$0.50          & 73.12$\pm$0.20          & 91.79$\pm$0.18          & 47.14$\pm$0.79          & 69.47$\pm$0.47          & \textbf{92.84$\pm$0.25} & 74.30$\pm$0.83          & 83.57$\pm$0.52          \\ \hline
    \multicolumn{1}{|l|}{KDDE(CDAN)}                                            & 90.81$\pm$0.67          & 53.40$\pm$1.96          & 72.10$\pm$1.31          & 89.96$\pm$0.45          & \textbf{49.73$\pm$0.83} & \textbf{69.84$\pm$0.33} & 92.14$\pm$0.31          & \textbf{75.23$\pm$0.42} & \textbf{83.68$\pm$0.33} \\ \hline
    \multirow{2}{*}{\textbf{Model}}                       & \multicolumn{3}{c|}{R$\rightarrow$A}                                        & \multicolumn{3}{c|}{R$\rightarrow$C}                                        & \multicolumn{3}{c|}{R$\rightarrow$P}                                        \\ \cline{2-10} 
                                                          & R                       & A                       & R+A                     & R                       & C                       & R+C                     & R                       & P                       & R+P                     \\ \hline
    \multicolumn{1}{|l|}{ResNet-50}                                             & 84.05$\pm$0.07          & 63.73$\pm$0.49          & 73.89$\pm$0.22          & \textbf{84.05$\pm$0.07} & 49.47$\pm$1.08          & 66.76$\pm$0.51          & \textbf{84.05$\pm$0.07} & 76.14$\pm$0.74          & 80.09$\pm$0.36          \\ \hline
    \multicolumn{1}{|l|}{DDC}                                                   & 84.49$\pm$0.26          & 64.52$\pm$0.50          & 74.50$\pm$0.30          & 83.21$\pm$0.28          & 53.23$\pm$0.87          & 68.22$\pm$0.54          & 84.02$\pm$0.33          & 77.79$\pm$0.42          & 80.91$\pm$0.24          \\ \hline
    \multicolumn{1}{|l|}{DANN}                                                  & 83.76$\pm$0.41          & 65.33$\pm$0.33          & 74.55$\pm$0.13          & 82.00$\pm$0.69          & 55.07$\pm$0.96          & 68.54$\pm$0.78          & 82.83$\pm$0.43          & 78.02$\pm$0.08          & 80.43$\pm$0.23          \\ \hline
    \multicolumn{1}{|l|}{DAAN}                                                  & 84.59$\pm$0.30          & 64.22$\pm$0.62          & 74.41$\pm$0.40          & 83.07$\pm$0.19          & 52.59$\pm$0.65          & 67.83$\pm$0.25          & 83.80$\pm$0.25          & 77.79$\pm$0.42          & 80.80$\pm$0.30          \\ \hline
    \multicolumn{1}{|l|}{CDAN}                                                  & 82.29$\pm$0.90          & 64.19$\pm$0.51          & 73.24$\pm$0.70          & 80.12$\pm$0.38          & 54.82$\pm$1.38          & 67.47$\pm$0.68          & 82.45$\pm$0.22          & 80.12$\pm$0.37          & 81.29$\pm$0.25          \\ \hline
    \multicolumn{1}{|l|}{KDDE(DDC)}                                             & \textbf{84.65$\pm$0.30} & \textbf{65.22$\pm$0.66} & \textbf{74.94$\pm$0.47} & 83.33$\pm$0.08          & 54.03$\pm$0.33          & \textbf{68.68$\pm$0.19} & 83.91$\pm$0.16          & 79.27$\pm$0.88          & 81.59$\pm$0.52          \\ \hline
    \multicolumn{1}{|l|}{KDDE(CDAN)}                                            & 82.94$\pm$0.31          & 64.55$\pm$0.90          & 73.74$\pm$0.59          & 80.25$\pm$1.33          & \textbf{56.73$\pm$1.02} & 68.49$\pm$1.04          & 82.79$\pm$0.56          & \textbf{80.79$\pm$0.73} & \textbf{81.79$\pm$0.58} \\ \hline
    \end{tabular}

    }
\end{table}

\section{Conclusions} \label{sec:conc}

We have defined a new task termed unsupervised domain expansion (UDE). Accordingly, two benchmark datasets, \ie Office-Home and DomainNet, have been re-purposed for the task. Our evaluation about four present-day domain adaptation models, either metric-based or adversarial, shows that their gain on the target domain is obtained at the cost of affecting their classification ability on the source domain. The proposed KDDE model effectively reduces such cost, and is found to be effective for both the new task and the traditional unsupervised domain adaptation task.

\medskip

\textbf{Acknowledgments}. This research was supported in part by National Natural Science
Foundation of China (No. 61672523), Beijing Natural Science Foundation (No. 4202033), the Fundamental Research Funds for the Central Universities and the Research Funds of Renmin University of China (No. 18XNLG19), and the Pharmaceutical Collaborative Innovation Research Project of Beijing Science and Technology Commission (No. Z191100007719002).

\bibliographystyle{ACM-Reference-Format}
\bibliography{da}

\appendix








\end{document}